\documentclass[manuscript]{acmart}
\AtBeginDocument{%
  }

\setcopyright{acmlicensed}
\copyrightyear{2025}
\acmYear{2025}
\acmDOI{XXXXXXX.XXXXXXX}

\usepackage{multirow}
\usepackage{threeparttable}
\DeclareCaptionLabelFormat{appendixtable}{Appendix Table~#2}
\begin{document}

%%
%% The "title" command has an optional parameter,
%% allowing the author to define a "short title" to be used in page headers.
\title{Label-Efficient Deep Learning in Medical Image Analysis: Challenges and Future Directions}

%%
%% The "author" command and its associated commands are used to define
%% the authors and their affiliations.
%% Of note is the shared affiliation of the first two authors, and the
%% "authornote" and "authornotemark" commands
%% used to denote shared contribution to the research.
\author{Cheng Jin}
\authornote{Both authors contributed equally to this research.}
\email{cheng.jin@connect.ust.hk}
\orcid{0000-0002-3522-3592}
\author{Zhengrui Guo}
\authornotemark[1]
\email{zguobc@connect.ust.hk}
\orcid{0009-0006-9920-0978}
\affiliation{%
  \institution{The Hong Kong University of Science and Technology}
  \country{Hong Kong SAR}
}

\author{Yi Lin}
\affiliation{
  \institution{Weill Cornell Medicine, New York, NY}
  \country{USA}
}
\orcid{0000-0002-7635-2518}
\email{yil4033@med.cornell.edu}

\author{Luyang Luo}
\affiliation{
  \institution{Harvard University, Cambridge, MA}
  \country{USA}
}
\orcid{0000-0002-7485-4151}
\email{luyang_luo@hms.harvard.edu}

\author{Hao Chen}
\authornote{Corresponding author.}
\affiliation{
 \institution{The Hong Kong University of Science and Technology}
 \country{Hong Kong SAR}
}
\orcid{0000-0002-8400-3780}
\email{jhc@cse.ust.hk}

\thanks{This work was supported by the National Natural Science Foundation of China (No. 62202403), Hong Kong Innovation and Technology Commission (Project No. MHP/002/22 and ITCPD/17-9), and Research Grants Council of the Hong Kong Special Administrative Region, China (Project No. R6003-22 and C4024-22GF).}

%%
%% By default, the full list of authors will be used in the page
%% headers. Often, this list is too long, and will overlap
%% other information printed in the page headers. This command allows
%% the author to define a more concise list
%% of authors' names for this purpose.
\renewcommand{\shortauthors}{Jin et al.}

%%
%% The abstract is a short summary of the work to be presented in the
%% article.
\begin{abstract}
Deep learning has significantly advanced medical imaging analysis (MIA), achieving state-of-the-art performance across diverse clinical tasks. However, its success largely depends on large-scale, high-quality labeled datasets, which are costly and time-consuming to obtain due to the need for expert annotation. To mitigate this limitation, label-efficient deep learning methods have emerged to improve model performance under limited supervision by leveraging labeled, unlabeled, and weakly labeled data. In this survey, we systematically review over 350 peer-reviewed studies and present a comprehensive taxonomy of label-efficient learning methods in MIA. These methods are categorized into four labeling paradigms: \textit{no label}, \textit{insufficient label}, \textit{inexact label}, and \textit{label refinement}. For each category, we analyze representative techniques across imaging modalities and clinical applications, highlighting shared methodological principles and task-specific adaptations. We also examine the growing role of health foundation models (HFMs) in enabling label-efficient learning through large-scale pre-training and transfer learning, enhancing the use of limited annotations in downstream tasks. Finally, we identify current challenges and future directions to facilitate the translation of label-efficient learning from research promise to everyday clinical care.
\end{abstract}

%%
%% The code below is generated by the tool at http://dl.acm.org/ccs.cfm.
%% Please copy and paste the code instead of the example below.
%%


%%
%% Keywords. The author(s) should pick words that accurately describe
%% the work being presented. Separate the keywords with commas.

\received{7 March 2025}

%% This command processes the author and affiliation and title
%% information and builds the first part of the formatted document.
\maketitle

\section{Introduction}
Deep learning (DL) has revolutionized medical image analysis (MIA), significantly improving the efficiency and accuracy of disease detection, diagnosis, and treatment \cite{de2018clinically, cao2023large, vorontsov2024foundation}. By providing a data-driven framework for interpreting large and diverse medical image datasets, DL models have achieved unprecedented performance.
% transition, importance
Despite these advancements, the success of DL models remains heavily dependent on large volumes of precisely annotated data, which are costly and time-consuming to obtain due to the need for expert input \cite{yu2021convolutional}. This growing demand for annotation contrasts sharply with the limited availability of medical experts \cite{rosenkrantz2016us, lu2020national}, creating a widening gap between the increasing volume of medical images and the capacity to label them. Reducing annotation costs, accelerating annotation workflows, and alleviating the burden on annotators have thus become critical challenges in DL-based MIA.

\begin{figure}[htbp]
\centering
\includegraphics[width=\textwidth]{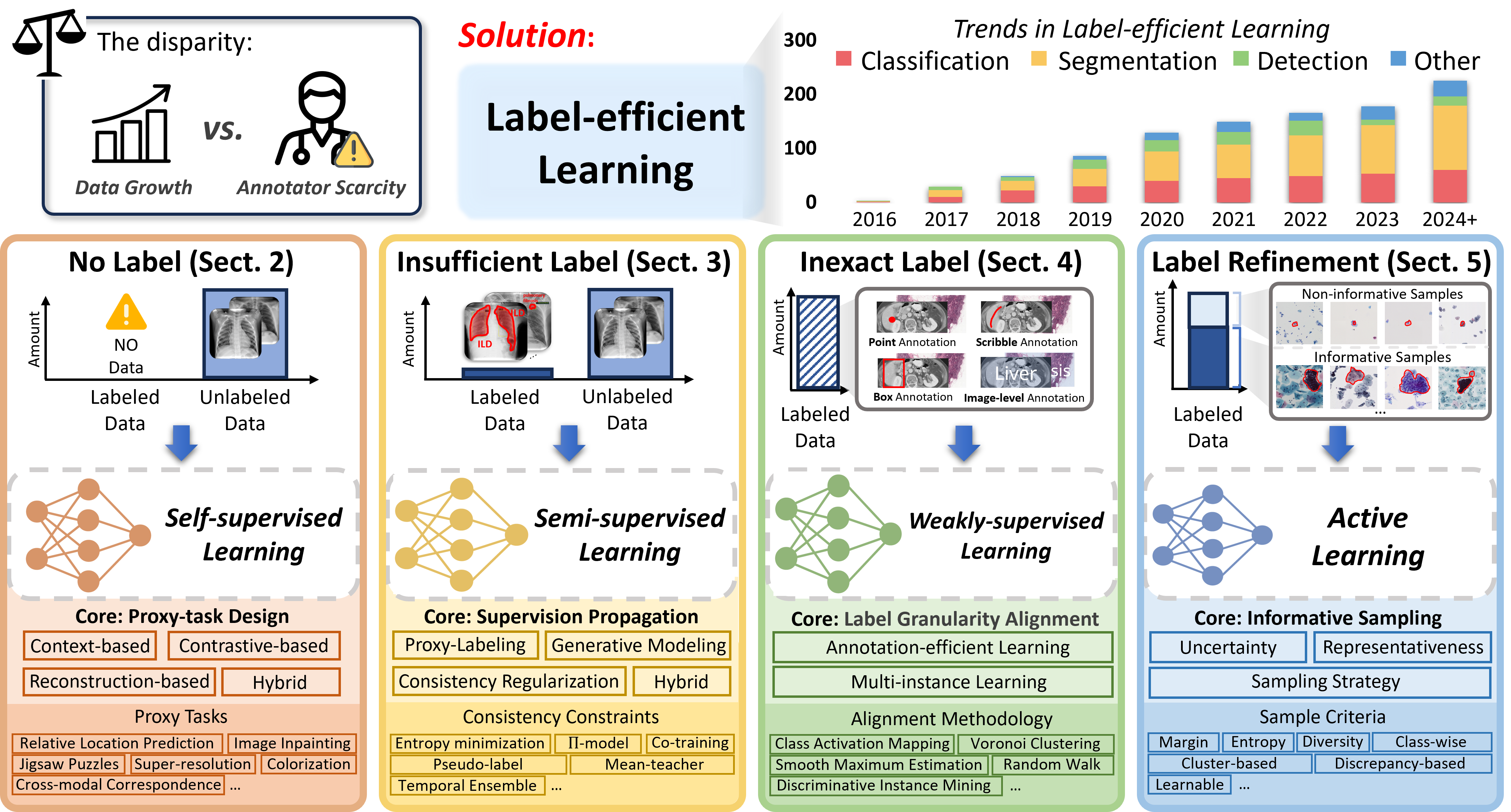}        
\caption{Overview of this survey. This survey categorizes approaches based on four labeling scenarios: No label (Section \ref{sec:ssl}), insufficient label (Section \ref{sec:semi}), inexact label (Section \ref{sec:mil}), and label refinement (Section \ref{sec:al}). This figure illustrates the disparity between data growth and annotator scarcity, the core techniques employed in each scenario, and trends in label-efficient learning applications. Detailed survey scope can be referred to Appendix \ref{appendix1}.}
	\label{fig_class}
\end{figure}

To address the annotation bottleneck in medical imaging, researchers have proposed a variety of learning paradigms, including self-supervised, semi-supervised, weakly supervised, and active learning. These approaches are designed to handle scenarios where annotations are missing, limited, imprecise, or require refinement. By leveraging different levels of supervision, ranging from pixel-level labels to weaker forms such as points, scribbles, bounding boxes, or even unlabeled data, they provide flexibility across diverse labeling conditions. The emergence of health foundation models (HFMs) has further strengthened these strategies by pretraining on large-scale medical datasets to extract generalizable features. These features can be effectively transferred to downstream tasks such as classification, segmentation, or detection, which improves performance and reduces the need for extensive labeled data during fine-tuning \cite{he2024foundation}. In this paper, we refer to the full spectrum of these methods, both traditional and HFM-based, as \textbf{label-efficient learning}.
As illustrated in Fig. \ref{fig_class}, label-efficient learning methods have rapidly expanded in recent years. High-level tasks such as classification, segmentation, and detection remain their primary focus, while applications to low-level tasks, including denoising, image registration, and super-resolution, are also gaining momentum. This growing versatility underscores the increasing impact of label-efficient learning across the MIA pipeline, supporting its broader integration into both research and clinical workflows.

Several surveys have previously addressed label-efficient learning in MIA, each offering valuable insights yet exhibiting certain limitations. Cheplygina et al. \cite{cheplygina2019not} introduced the term ``not-so-supervised'' learning and categorized methods into supervised, semi-supervised, multi-instance, and transfer learning. However, their work focused on theoretical concepts and lacked practical relevance to clinical MIA applications. Budd et al. \cite{budd2021survey} emphasized human-in-the-loop strategies, while Wang et al. \cite{wang2024comprehensive} provided a more recent survey, but covered only a subset of label-efficient methods, limiting its scope. In contrast, our taxonomy is organized around specific annotation scenarios, offering a more intuitive and application-oriented framework. Furthermore, we analyze how HFMs are reshaping traditional paradigms within each scenario, highlighting open research questions that merit future exploration, as shown in Fig. \ref{fig_class}.

To provide a comprehensive overview of label-efficient learning in MIA, we review over 350 peer-reviewed studies and categorize them into four annotation scenarios: \textit{no label}, where data lacks annotations; \textit{insufficient label}, where labeled data is limited; \textit{inexact label}, where annotations are noisy or coarse; and \textit{label refinement}, where existing labels require improvement. To the best of our knowledge, this is the first extensive review to systematically cover all four scenarios. For each, we define the core challenges, provide essential background, and examine the role of HFMs in enhancing performance. Our analysis not only synthesizes recent progress but also identifies key limitations and outlines future research directions, offering a roadmap for advancing label-efficient learning in MIA. The remainder of this paper is organized as follows. Sections \ref{sec:ssl}--\ref{sec:al} introduce the four annotation scenarios: \textit{no label} in Section \ref{sec:ssl}, \textit{insufficient label} in Section \ref{sec:semi}, \textit{inexact label} in Section \ref{sec:weakly}, and \textit{label refinement} in Section \ref{sec:al}. Section \ref{sec:cnfd} discusses current challenges and explores potential solutions and research opportunities. Finally, we conclude this survey in Section \ref{sec:con}.
\section{No Label} \label{sec:ssl}
\begin{figure}[htbp]
	\centering
\includegraphics[width=0.95\textwidth]{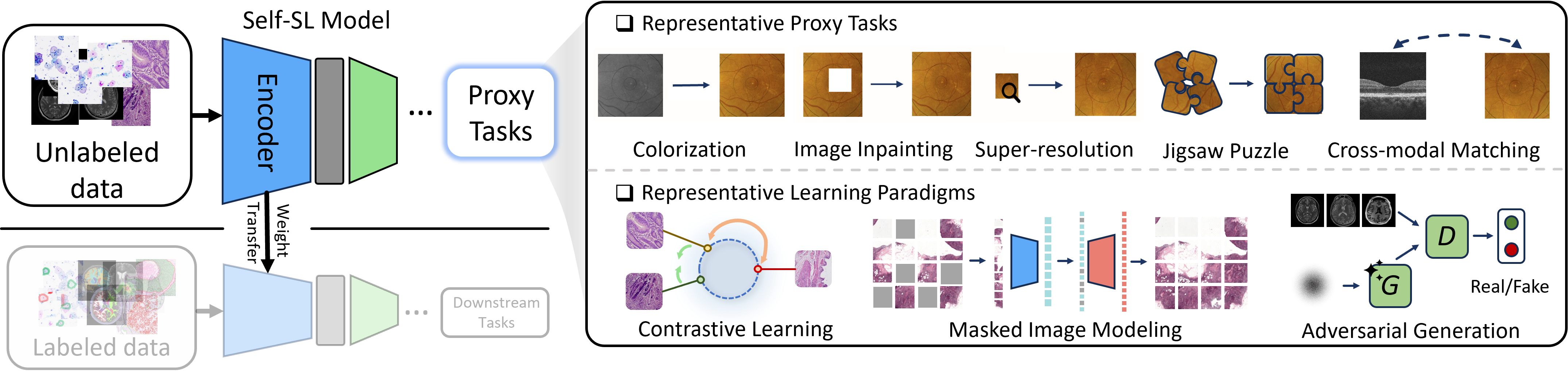}
	\caption{Overview of self-supervised learning (Self-SL) paradigm which addresses the \textit{no label} scenario. Self-SL aims to learn a pre-trained model by developing various proxy tasks based solely on unlabeled data. Then the pre-trained model can be fine-tuned on different downstream tasks with labeled datasets.}
	\label{fig_self_schematic}
\end{figure}

The \textit{no label} scenario represents one of the most challenging yet prevalent settings in MIA, where large volumes of imaging data are available but no annotations exist. This commonly arises in applications involving novel modalities, rare diseases, or legacy datasets lacking corresponding labels. In the absence of ground truth, conventional supervised learning becomes inapplicable. To address this, \textbf{self-supervised learning (Self-SL)} has emerged as an effective solution. Self-SL derives supervisory signals directly from the data itself by designing pretext tasks that exploit the inherent structural and semantic patterns within medical images. As illustrated in Fig.~\ref{fig_self_schematic}, Self-SL enables models to learn meaningful and transferable representations from unlabeled data through automatically generated supervision signals. Typically, the workflow involves first pretraining a model on large-scale unlabeled data using self-supervised objectives, followed by fine-tuning on a labeled dataset. This strategy not only facilitates data-specific feature learning but also helps reduce overfitting by leveraging the abundance of unlabeled data. Moreover, recent advances have scaled up both the size of datasets and the capacity of models, giving rise to health foundation models (HFMs). These models have demonstrated strong diagnostic and prognostic performance, along with robust generalization across diverse medical imaging tasks. Based on the design of proxy tasks, we categorize Self-SL methods in MIA as \textbf{reconstruction-based}, \textbf{context-based}, and \textbf{contrastive-based} approaches. Representative approaches within each category are summarized in Appendix Table~\ref{tab:self}. 

\subsection{Reconstruction-Based Methods}

\textbf{Reconstruction-based methods} aim to learn semantic features by reconstructing input data, enabling representation learning without manual annotations. These methods encompass tasks such as super-resolution \cite{zhao2020smore,li2021single}, inpainting \cite{zhao2021anomaly}, colorization \cite{abbet2020divide}, and multi-modal reconstruction \cite{hervella2018retinal,cao2020auto}.

A basic form of reconstruction was introduced by Li \textit{et al.} \cite{li2020sacnn}, who trained an auto-encoder to reconstruct normal-dose CT images by minimizing mean squared error (MSE) loss. The encoder, pre-trained via self-supervision, was later reused for downstream tasks. However, reliance on pixel-wise losses may bias models toward low-level appearance features while overlooking structural semantics \cite{abbet2020divide}, prompting the development of more informative proxy tasks.

Super-resolution tasks generate high-resolution outputs from low-resolution inputs, forcing the model to learn fine-grained semantic structures. Zhao \textit{et al.} \cite{zhao2020smore} proposed an anti-aliasing super-resolution method for MRIs, while Li \textit{et al.} \cite{li2023generic} leveraged frequency information for fundus image enhancement. GAN-based super-resolution has also been applied to gigapixel histopathology WSIs \cite{li2021single}. These super-resolution approaches are particularly effective in domains where high-resolution details are critical for diagnosis, and they encourage the model to focus on subtle anatomical structures. However, they may also lead to overfitting to texture or noise if not properly regularized, and their effectiveness can be limited in cases where semantic content is not directly tied to resolution.

Colorization tasks predict RGB images from grayscale inputs, encouraging the model to capture structural and contextual cues. Abbet \textit{et al.} \cite{abbet2020divide} applied this to colorectal cancer survival analysis by converting images into hematoxylin and eosin channels. Fan \textit{et al.} \cite{fan2023cancerself} extended this approach with a cross-channel task that reconstructs lightness from color channels. Compared to pure pixel reconstruction, colorization requires the model to infer semantic correspondences between structure and color, thus promoting higher-level feature learning. Nonetheless, the colorization task may introduce ambiguity, as multiple plausible color mappings can exist for the same structure, potentially limiting the precision of the learned representations.

Inpainting tasks require models to restore missing regions, promoting understanding of object continuity and context. Zhao \textit{et al.} \cite{zhao2021anomaly} applied this strategy to OCT and chest X-rays, showing improved anatomical feature learning. Inpainting is especially useful for encouraging global context reasoning, as the model must integrate information from surrounding regions to plausibly reconstruct missing parts. However, if the occluded regions are too small or too easily inferred from context, the task may become trivial and fail to drive the model towards learning meaningful semantics.

Multi-modal reconstruction tasks aim to reconstruct one modality from another, facilitating joint representation learning. Hervella \textit{et al.} \cite{hervella2018retinal} combined retinography and angiography for retinal analysis, while Cao \textit{et al.} \cite{cao2020auto} proposed a collaborative learning framework using auto-encoders and GANs to synthesize missing modalities. These approaches are well-suited for leveraging complementary information across imaging modalities, which is common in clinical practice. Yet, the challenge remains in aligning heterogeneous data distributions and ensuring that the shared representations capture clinically relevant features rather than modality-specific artifacts.

Recent advances in diffusion models \cite{ho2020denoising,yang2022diffusion,rombach2022high} have further improved self-supervised reconstruction. These models, characterized by stability and generative fidelity, have been applied to medical imaging by Korkmaz \textit{et al.} \cite{korkmaz2023self}, Cho \textit{et al.} \cite{cho2023improved}, and Purma \textit{et al.} \cite{purma2024genselfdiff}, enabling robust feature learning from complex anatomical data. While diffusion models offer superior generative performance and can model complex data distributions, they typically require significant computational resources and their interpretability in clinical contexts remains an open question.

In summary, reconstruction-based self-supervised methods provide a flexible framework for learning rich representations from unlabeled medical images. Each proxy task emphasizes different aspects of image semantics—whether fine-grained structure, global context, or cross-modal relationships. However, these methods also face challenges, such as balancing low-level appearance with high-level semantics, handling ambiguous or ill-posed reconstruction targets, and ensuring clinical relevance of learned features. Going forward, integrating more task-aware proxy objectives and leveraging advances in generative modeling hold promise for further improving the effectiveness and applicability of reconstruction-based self-supervised learning in medical imaging.

\subsection{Context-Based Methods}
While reconstruction-based methods focus on pixel-level restoration, \textbf{context-based methods} take a different approach by exploiting the spatial and structural relationships within medical images. Researchers have explored novel predictive tasks for specific MIA tasks by training the network for prediction of the output class or localization of objects with the original image as the supervision signal \cite{bai2019self,spitzer2018improving,srinidhi2022self}. Bai \textit{et al.} \cite{bai2019self} propose a proxy task to predict the anatomical positions from cardiac chamber view planes by applying an encoder-decoder structure. This proxy task properly employs the chamber view plane information, which is available from cardiac MR scans easily. 
Zheng \textit{et al.} \cite{zheng2023msvrl} enhance medical image segmentation through multi-scale consistency objectives for finer-grained representation across different target scales. For 3D medical images, He \textit{et al.} \cite{he2023geometric} introduce Geometric Visual Similarity Learning, incorporating topological invariance prior to ensure consistent representation of semantic regions. Meanwhile, Srinidhi \textit{et al.} \cite{srinidhi2022self} propose Resolution Sequence Prediction for whole slide images (WSIs), training networks to predict the order of multi-resolution patches, thereby capturing both contextual structure at lower magnifications and local details at higher magnifications.

Other efforts have been made to explore the spatial context structure of input data, such as the order of different patches constituting an image, or the relative position of several patches in the same image, which can provide useful semantic features for the network. 
% Concise version
Chen \textit{et al.} \cite{chen2019self} introduce context restoration, where patch positions are randomly switched and then restored, enabling straightforward semantic feature learning. Li \textit{et al.} \cite{li2021rotation} employ rotation angle prediction with augmented retinal images, encouraging the model to predict rotation angles while clustering similar features. Advanced spatial reasoning tasks such as jigsaw puzzles and Rubik's Cube have gained traction in medical imaging. Taleb \textit{et al.} \cite{taleb2021multimodal} enhance Jigsaw Puzzles with multi-modal data, requiring models to restore original images from out-of-order patches of different modalities. For 3D medical data, Zhuang \textit{et al.} \cite{zhuang2019selfsupervised} and Tao \textit{et al.} \cite{tao2020revisiting} adapt Rubik's Cube by dividing volumes into cube grids, applying random rotations, and training networks to recover the volume.

However, for WSI images, common proxy tasks such as prediction of the rotation or relative position of objects may only provide minor improvements to the model in histopathology due to the lack of a sense of global orientation in WSIs \cite{graham2020dense,koohbanani2021self}. To address these limitations, Koohbanani \textit{et al.} \cite{koohbanani2021self} proposes proxy tasks targeted at histopathology, namely, magnification prediction, solving the magnification puzzle, and hematoxylin channel prediction. In this way, their model can significantly integrate and learn the contextual, multi-resolution, and semantic features inside the WSIs.

In summary, context-based self-supervised methods provide a versatile framework for capturing semantic and structural information in medical images. Their success largely depends on the careful design of proxy tasks that align with the intrinsic properties of the data and the requirements of downstream applications. Nonetheless, challenges remain in selecting context cues that are both informative and robust across diverse imaging modalities. Future work may focus on dynamic or adaptive task selection, as well as integrating multiple context signals to further enhance feature learning in MIA.

\subsection{Contrastive-Based Methods}
Beyond the pixel-space focus of reconstruction approaches and the spatial reasoning of context-based methods, \textbf{contrastive-based methods} operate on the principle that representations of different views of the same image should be similar, while those of different images should be distinguishable. Several high-performance contrastive learning algorithms originally developed for natural images, such as SimCLR \cite{chen2020simple} and BYOL \cite{grill2020bootstrap}, have been successfully adapted to medical image analysis \cite{azizi2021big,wang2021transpath}. Azizi \textit{et al.} \cite{azizi2021big} developed multi-instance contrastive learning (MICLe), extending SimCLR by minimizing disagreement between views from multiple images of the same patient, creating richer positive pair relationships. In parallel, Wang \textit{et al.} \cite{wang2021transpath} applied the BYOL architecture to histopathology image classification, making a substantial contribution by assembling the largest WSI dataset for Self-SL pre-training at the time—comprising 2.7 million patches from 32,529 WSIs spanning over 25 anatomic sites and 32 cancer subtypes.

Large-scale dataset utilization has emerged as a critical trend in contrastive-based methods. Ghesu \textit{et al.} \cite{ghesu2022self} developed a contrastive learning and online clustering algorithm leveraging over 100 million radiography, CT, MRI, and ultrasound images. This extensive pre-training yielded significant improvements in both performance and convergence rates compared to previous state-of-the-art approaches. Similarly, Nguyen \textit{et al.} \cite{nguyen2023lvm} integrated over 1.3 million multi-modal images from 55 publicly available datasets to enhance representation learning. Beyond multi-view approaches, Jiang \textit{et al.} \cite{jiang2023anatomical} introduced a specialized contrastive objective for learning anatomically invariant features, designed to exploit inherent similarities in anatomical structures across diverse medical imaging volumes.

Subsequent research has further refined contrastive learning by incorporating both global and local contrast for more comprehensive representation learning, typically using InfoNCE loss \cite{oord2018representation}.

Yan \textit{et al.} \cite{yan2022sam} apply this at the pixel level to generate embeddings that accurately describe anatomical locations, creating representations at both global and local scales. Building on this multi-level approach, Liu \textit{et al.} \cite{liu2023hierarchical} develop hierarchical contrastive learning for intra-oral mesh scans, capturing unsupervised representations across point-level, region-level, and cross-level interactions.

Compared to reconstruction- and context-based methods, contrastive learning is less reliant on proxy task design and better suited to leveraging large unlabeled datasets, often resulting in more robust and transferable representations. However, it faces challenges such as constructing meaningful positive/negative pairs and mitigating spurious correlations, especially in settings with limited patient diversity or multi-modal data. Looking ahead, advances in pair mining and domain-adaptive objectives are expected to further improve the effectiveness and clinical relevance of contrastive-based self-supervised learning in medical imaging.

\subsection{Discussion}
Self-SL methods aim to leverage unlabeled data to learn rich, transferable representations by designing effective proxy tasks. While many existing approaches directly adapt proxy tasks from natural image domains, the unique properties of medical images—such as CT, WSI, and MRI—necessitate tailored proxy task designs that account for domain-specific semantics and structures. 
% Newly Added
The choice of Self-SL strategy often depends on the imaging modality: reconstruction-based methods generally perform well for modalities with clear anatomical structure (like CT and MRI), contrastive-based approaches excel with high-dimensional histopathology data that benefits from instance discrimination, while context-based methods are particularly effective when spatial relationships are diagnostically significant. Researchers have also explored hybrid frameworks that combine multiple types of Self-SL tasks to capture diverse aspects of medical data (see \cite{tang2022self,zhou2021preservational,haghighi2020learning,yang2022cs} and references therein).

Looking forward, the design of proxy tasks that incorporate modality-specific information and multi-modal integration remains a promising direction, enabling models to disentangle and capture complementary features from different imaging sources. Large vision-language pre-trained models~\cite{zhou2023advancing,zhou2022generalized,park2023self} have recently shown great potential in chest X-ray analysis, attracting increasing research interest. Moreover, the field is witnessing a rapid scaling of both data and model size, exemplified by the emergence of HFMs~\cite{ma2024segment,lu2024visual,chen2024towards,vorontsov2024foundation}. For instance, MedSAM~\cite{ma2024segment} adapts the Segment Anything Model to medical segmentation by fine-tuning on 1.57 million image-mask pairs across 10 modalities, while CONCH~\cite{lu2024visual} extends contrastive learning to whole-slide images with over 1 million image-caption pairs. Similarly, large-scale pre-training efforts such as UNI~\cite{chen2024towards} and Virchow~\cite{vorontsov2024foundation} demonstrate the power of scaling, using hundreds of thousands to over a million whole-slide images to learn generalizable representations. These trends highlight the increasing benefits of leveraging larger and more diverse datasets, as well as more powerful model architectures, to capture the complex patterns inherent to medical imaging.
\section{Insufficient Label} \label{sec:semi}
The \textit{insufficient label} scenario arises when only a small portion of available medical imaging data is annotated, while the majority remains unlabeled. This scenario is prevalent in clinical settings, where expert annotations are costly and time-consuming, yet raw images are readily accessible. In such cases, supervised learning alone proves inadequate due to limited labeled data. 
\begin{figure}[htbp]
\centering
\includegraphics[width=0.9\textwidth]{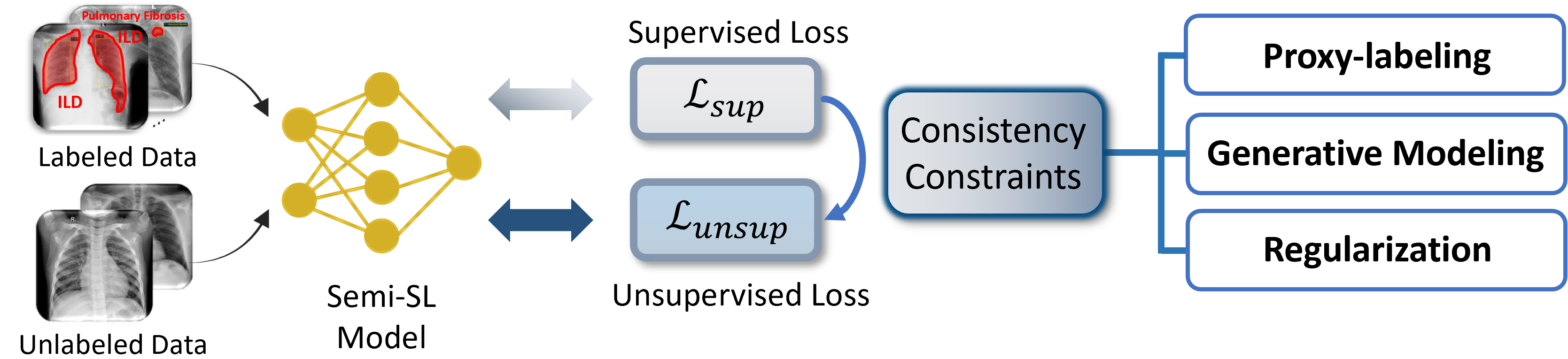}
	\caption{Overview of Semi-supervised learning (Semi-SL) paradigm which addresses the \textit{insufficient label} scenario. Semi-SL leverages a small amount of labeled data and a large amount of unlabeled data to jointly train a model. The goal is to ensure prediction consistency between labeled and unlabeled examples, typically by optimizing a combination of supervised ($\mathcal{L}_{sup}$) and unsupervised ($\mathcal{L}_{unsup}$) loss functions.} \label{fig_semi_schematic}
\end{figure}
As illustrated in Fig.~\ref{fig_semi_schematic}, \textbf{semi-supervised learning (Semi-SL)} addresses this challenge by leveraging both labeled and unlabeled data during training. A core principle of Semi-SL is supervision propagation, which assumes that unlabeled data should have predictions consistent with labeled data. In this way, Semi-SL enables better generalization while reducing the need for manual annotation.

In the following, we categorize existing Semi-SL methods in MIA how each method enforces or approximates supervision consistency between labeled and unlabeled data into three categories: \textbf{proxy-labeling}, \textbf{generative modeling}, and \textbf{regularization}. Representative works are summarized in Appendix Tab.~\ref{tab:semi}.

\subsection{Proxy-labeling Methods}
\textbf{Proxy-labeling methods} utilize the idea of supervision consistency propagation by assigning pseudo labels to unlabeled samples and incorporate high-confidence examples into the training process through an iterative approach. These methods can be divided into two principal subcategories: \textit{Self-training methods} and \textit{multi-view learning methods}.

\subsubsection{Self-training Methods}

\textit{Self-training methods} operate through the bootstrapping mechanism. Initially, a prediction function $f_{\theta}$ with parameters $\theta$ is trained using available labeled data samples $x\in X_L$. Subsequently, this trained model generates predictions for unlabeled data samples $x\in X_U$. A confidence threshold $\tau$ is established, and sample-label pairs $(x, \mathrm{argmax}{f_{\theta}(x)})$ whose prediction confidence exceeds $\tau$ are added to the labeled dataset $X_L$. This augmented labeled dataset is then used to retrain the prediction function, creating an iterative cycle that continues until the model can no longer make sufficiently confident predictions on remaining unlabeled data.

Entropy minimization \cite{grandvalet2004semi} represents a foundational approach in this category, regularizing models based on the low-density assumption by encouraging low-entropy predictions for unlabeled data. Building on this concept, \textbf{Pseudo-label} \cite{lee2013pseudo} provides a straightforward yet effective self-training mechanism that applies entropy minimization principles in prediction space. While labeled samples undergo supervised training, unlabeled data receive labels corresponding to the model's most confident predictions. In medical image analysis, Pseudo-label has been widely employed as an auxiliary component to enhance model performance across various applications \cite{fan2020inf,zhang2022boostmis,chaitanya2023local}.

A significant challenge with proxy labels is their inherent noise and potential deviation from ground truth. To address this limitation, researchers have developed quality assurance mechanisms including uncertainty-aware confidence evaluation \cite{wang2021semiself}, conditional random field-based proxy label refinement \cite{bai2017semi}, and adversarial training-based methods \cite{zhou2019collaborative}. These approaches help ensure that proxy labels provide reliable supervisory signals during training. Pseudo-label has also been used in MIA to refine a given annotation with the assistance of unlabeled data. Qu \textit{et al.} \cite{qu2020weakly} introduce pseudo-label into nuclei segmentation and design an iterative learning algorithm to refine the background of weakly labeled images where only nuclei are annotated, leaving large areas ignored. Similar ideas can also be seen in \cite{nie2018asdnet}.

\subsubsection{Multi-view learning methods}
\textit{Multi-view learning methods} assume that each sample has two or multiple complementary views and features of the same sample extracted with different views are supposed to be consistent. Therefore, the key idea of multi-view learning methods is to train the model with multiple views of the sample or train multiple learners and minimize the disagreement between them, thus learning the underlying features of the data from multiple aspects. \textbf{Co-training} is a method that falls into this category. It assumes that data sample $x$ can be represented by two views, $\textbf{v}_1(x)$ and $\textbf{v}_{2}(x)$, and each of them are capable of solely training a good learner, respectively. Consequently, the two learners are set to make predictions of each view's unlabeled data, and iteratively choose the candidates with the highest confidence for the other model \cite{yang2021survey}. 
Another variation of multi-view learning methods is Tri-training \cite{zhou2005tri}, which is proposed to tackle the lack of multi-view data and mistaken labels of unlabeled data produced by self-training methods. Tri-training aims to learn three models from three different training sets obtained with bootstrap sampling. A deep learning version of Tri-training, i.e. Tri-Net, has been further proposed in \cite{dong2018tri}.

Co-training, or deep co-training, is dominant in multi-view learning in MIA, with a steady flow of publications \cite{zhao2019multi,zhou2019semi,xia2020uncertainty,wang2021selfco,fang2020dmnet,zeng2023pefat}. To conduct whole brain segmentation, Zhao \textit{et al.} \cite{zhao2019multi} implements co-training by obtaining different views of data with data augmentation. A similar idea can be seen for 3D medical image segmentation in \cite{xia2020uncertainty} and \cite{zhou2019semi}. These two works both utilize co-training by learning individual models from different views of 3D volumes such as the sagittal, coronal, and axial planes. Further works have been proposed to refine co-training. To produce reliable and confident predictions, Wang \textit{et al.} \cite{wang2021selfco} develops a self-paced learning strategy for co-training, forcing the network to start with the easier-to-segment regions and transition to the difficult areas gradually. Rather than discarding samples with low-quality pseudo-labels, Zeng \textit{et al.} \cite{zeng2023pefat} introduces a novel regularization approach, which focuses on extracting discriminative information from such samples by injecting adversarial noise at the feature level, thereby smoothing the decision boundary.
Meanwhile, to avoid the errors of different model components accumulating and causing deviation, Fang and Li \cite{fang2020dmnet} develop an end-to-end model called difference minimization network for medical image segmentation by conducting co-training with an encoder shared by two decoders.

\subsection{Generative Modeling Methods}
While proxy-labeling methods directly assign labels to unlabeled data, \textbf{generative modeling methods} realize supervision consistency propagation by assuming that both labeled and unlabeled data are sampled from a shared latent distribution. By learning this underlying distribution with the help of unlabeled data, these methods enable the model to transfer information across the entire dataset. The learned latent representation is then combined with supervised information from labeled examples to further improve performance.

% Concise version
\textbf{Generative adversarial networks (GANs)} effectively leverage both labeled and unlabeled data through a two-player minimax game between a generator $\mathcal{G}$ and discriminator $\mathcal{D}$ \cite{goodfellow2014generative}. In medical image analysis, several semi-supervised approaches incorporate unlabeled data during adversarial training. Chaitanya \textit{et al.} \cite{chaitanya2021semi} and Hou \textit{et al.} \cite{hou2022semi} utilize unlabeled samples to introduce greater variation in shape and intensity, enhancing model robustness. Zhou \textit{et al.} \cite{zhou2019collaborative} generate pseudo lesion masks for unlabeled data with quality facilitated by the discriminator. Other researchers modify the discriminator's objective beyond binary classification: Odena \textit{et al.} \cite{odena2016semi} extend it to predict $K$ classes plus an additional real/fake class, allowing unlabeled data to contribute to multi-class discrimination. This architecture has been successfully applied to retinal image synthesis \cite{kamran2021vtgan, diaz2019retinal, xie2023fundus}, glaucoma assessment \cite{diaz2019retinal}, chest X-ray classification \cite{madani2018semi}, and other medical imaging tasks \cite{hou2022semi}.

\textbf{Variational autoencoders (VAEs)} offer another effective approach for utilizing unlabeled data. Based on Bayesian inference theory \cite{kingma2013auto}, VAEs encode data into latent variables and reconstruct inputs by maximizing the variational lower bound. In medical image analysis, VAEs primarily learn feature similarities from large unlabeled datasets, creating well-constrained latent spaces that reduce dependence on labeled data \cite{sedai2017semi, wang2022rethinking}. Sedai \textit{et al.} \cite{sedai2017semi} proposed a dual-VAE framework for semi-supervised optic cup segmentation in retinal images, where one VAE learns data distribution from unlabeled data and transfers this knowledge to a second VAE performing segmentation with labeled data. Wang \textit{et al.} \cite{wang2022rethinking} adapted VAEs for 3D medical image segmentation by replacing the conventional mean vector and variance vector with a mean vector and covariance matrix, accounting for correlations between different slices of an input volume.

More recently, \textbf{diffusion models} \cite{ho2020denoising,yang2022diffusion,rombach2022high} have emerged as powerful alternatives in the generative Semi-SL domain. These models offer enhanced stability and sample quality through iterative denoising processes, showing potential in midline shift quantification \cite{gong2023diffusion} and medical image segmentation \cite{liu2024diffrect}. Their ability to model complex anatomical structures while enabling uncertainty quantification makes them particularly valuable for label-scarce scenarios.

\subsection{Regularization-based Methods}
In contrast to the explicit labeling approach of proxy methods and the distribution modeling of generative techniques, regularization-based methods enforce consistency through direct constraints on the model's behavior by assuming that the perturbation of data points does not change the prediction of the model, without requiring any label information. 

$\boldsymbol{\Pi}$\textbf{-model} \cite{sajjadi2016regularization} effectively implements consistency regularization by using a shared encoder to process differently augmented views of the same input and enforcing consistent predictions across these views, while incorporating label information to improve classifier performance. Li \textit{et al.} \cite{li2018semipi} achieved state-of-the-art skin lesion segmentation using this approach with only 300 labeled images, outperforming fully-supervised methods that required 2,000 labeled images. Similar consistency-based approaches appear in Bortsova \textit{et al.} \cite{bortsova2019semi}, who enforce prediction consistency across transformations for chest X-ray segmentation, and Meng \textit{et al.} \cite{meng2023dual}, who employ graph convolution networks to maintain regional and marginal consistency for semi-supervised optic disc and cup segmentation.

\textbf{Temporal ensembling} \cite{laine2016temporal} improves the $\Pi$-model's prediction stability by incorporating exponential moving averages, an approach widely adopted in medical image analysis \cite{cao2020uncertainty,gyawali2019semi,shi2020graph,luo2020deep}. For breast mass segmentation, Cao \textit{et al.} \cite{cao2020uncertainty} integrate uncertainty maps as guidance to ensure prediction reliability. Similarly, Luo \textit{et al.} \cite{luo2020deep} propose uncertainty-aware temporal ensembling for chest X-ray screening with partially labeled data. Gyawali \textit{et al.} \cite{gyawali2019semi} enhance the method by first using a VAE to extract disentangled latent space representations as stochastic embeddings, improving chest X-ray classification performance. A key characteristic of temporal ensembling is that each training sample's activation is updated only once per epoch.
 
\textbf{Mean teacher} \cite{tarvainen2017mean} applies exponentially moving average to model parameters rather than network activations, addressing the limitations of temporal ensembling and finding various applications in medical imaging \cite{li2020transformation,yu2019uncertainty,wang2020double,xu2023ambiguity,adiga2023anatomically}. Li \textit{et al.} \cite{li2020transformation} apply this approach to transformation-consistent medical image segmentation. Since teacher models can generate inaccurate targets for unlabeled data, Yu \textit{et al.} \cite{yu2019uncertainty} and Adiga \textit{et al.} \cite{adiga2023anatomically} incorporate uncertainty maps to ensure target reliability. Wang \textit{et al.} \cite{wang2020double} further propose a double-uncertainty-weighted method for left atrium and kidney segmentation, extending uncertainty from segmentation to feature level. Xu \textit{et al.} \cite{xu2023ambiguity} focus on selecting productive unsupervised consistency targets through an ambiguity-consensus mean-teacher model that better exploits complementary information from unlabeled data.

\subsection{Discussion}
Semi-supervised learning addresses the scarcity of labeled data by exploiting large amounts of unlabeled samples and enforcing supervision consistency across the dataset. 
% Newly Added
Data characteristics and task requirements should guide the choice of the Semi-SL strategy: proxy-labeling methods like self-training tend to perform well when high-confidence predictions can be reliably identified; multi-view learning approaches appear particularly suited for volumetric data where different perspectives provide complementary information; generative modeling shows promise with complex anatomical structures that benefit from learned prior distributions; while regularization-based methods often demonstrate robustness across diverse imaging modalities.
A persistent challenge in Semi-SL lies in the utilization of noisy or imperfect unlabeled data. The generation and selection of reliable pseudo labels are critical, as label noise can easily propagate through the training process and undermine model performance. Moreover, the theoretical understanding of how different Semi-SL techniques interact within hybrid systems remains limited, especially when dealing with heterogeneous data sources~\cite{wang2021deephybrid, zhang2022boostmis, wang2020focalmix, gyawali2020semi,miao2023caussl}. 

Meanwhile, the emergence of HFMs has significantly reshaped the landscape of semi-supervised learning in medical image analysis. As demonstrated by approaches like SemiSAM \cite{zhang2023semisam} and SemiSAM+ \cite{zhang2025semisam+}, these models introduce a paradigm shift from traditional model-centric Semi-SL methods focused on regularization strategies toward leveraging pre-trained knowledge to guide the learning process. Foundation models trained on large-scale datasets provide robust prior knowledge that helps specialist models learn more effectively with extremely limited labeled data—a scenario where conventional Semi-SL methods often struggle. This collaborative learning approach, where trainable specialist models interact with frozen foundation models, offers several advantages: it enhances performance in low-annotation regimes, provides more stable training due to knowledge transfer, and exhibits strong generalization capabilities across different medical imaging modalities and targets. As HFMs continue to evolve with improved architectures and more diverse training data, they will likely further transform Semi-SL in MIA, potentially reducing annotation requirements while increasing effectiveness and robustness.
\section{Inexact Label} \label{sec:weakly}
Obtaining precise, pixel-level annotations in MIA is often prohibitively expensive or time-consuming. In practice, clinicians frequently provide only incomplete or coarse annotations—such as sparse points or image-level labels—resulting in the so-called \textbf{inexact label} setting, where the available supervision does not match the granularity required for the target task. \textbf{Weakly supervised learning (WSL)} provides a general framework for addressing the challenges posed by inexact labels. The core issue in this context is label granularity alignment, which refers to bridging the gap between coarse-grained annotations and the fine-grained predictions demanded by clinical applications.
\begin{figure}[htbp]
	\centering
\includegraphics[width=\textwidth]{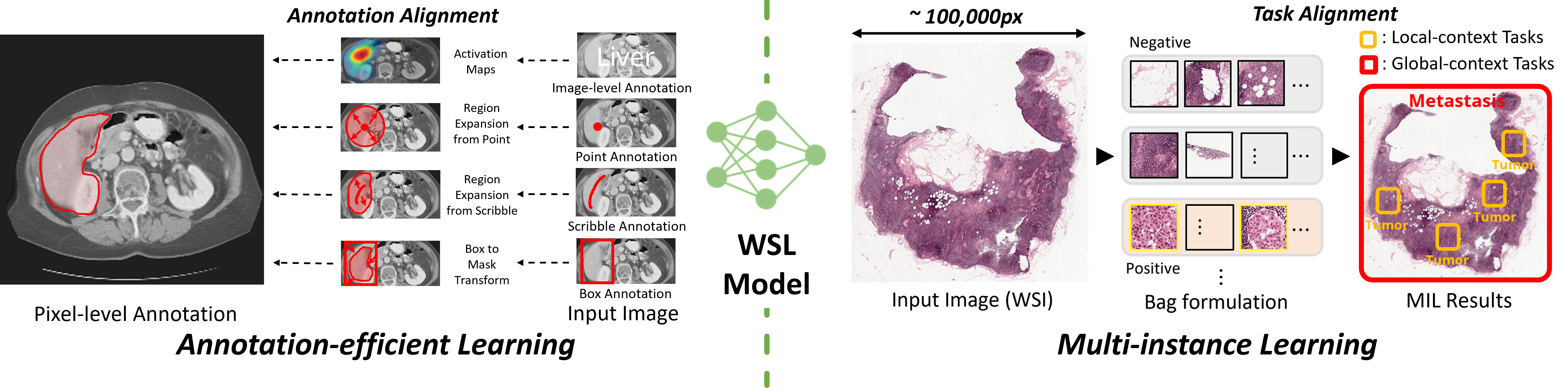}
	\caption{Overview of weakly-supervised learning paradigm which addresses the \textit{inexact label} scenario. On the annotation side, annotation-efficient learning strategy aims to maximize the utility of limited or partially annotated data while minimizing annotation burden. On the task side, multi-instance learning strategy aims to bridge the label granularity gap by inferring fine-grained predictions from coarse supervision.
    }
	\label{fig_wsl_schematic}
\end{figure}
Granularity alignment in WSL can be approached from two complementary perspectives. On the annotation side, \textbf{annotation-efficient learning} scheme aims to maximize the utility of limited or partially annotated data while minimizing annotation burden. These methods leverage both labeled and unlabeled data to enhance model performance with reduced annotation cost. On the task side, the challenge is to infer fine-grained information, such as pixel-level segmentation or lesion localization, from supervision that is only available at a coarser level, such as the whole image. \textbf{Multiple instance learning (MIL)} provides a representative learning scheme for this scenario by associating global image-level labels with collections of instances, such as patches or regions, and enabling the model to identify relevant instances based on weak supervision.

In the following, we elaborate on the principles and representative applications of these strategies, with a focus on how label granularity alignment enables robust learning from inexact labels in MIA.

\subsection{Annotation-Efficient Learning}\label{sec:anno}
To address label granularity alignment from the annotation perspective, \textbf{annotation-efficient learning} leverages deep learning techniques with partially labeled data to enhance labeling efficiency for dense predictions. A strategic approach to boost annotation efficiency is to utilize markings other than complete dense annotations. 
While there may be some overlap with other methodologies, annotation-efficient learning methods are specifically tailored to harness the unique attributes of various annotation forms to improve efficiency and bridge the granularity disparity between annotation and prediction.
In this section, we review representative annotation-efficient learning strategies that address label granularity alignment through a coarse-to-fine approach, focusing on techniques related to \textbf{image-level}, \textbf{point}, \textbf{scribble}, and \textbf{box} annotations. 
Appendix Tab.~\ref{tab:anno} provides an overview of notable publications in this domain.

\subsubsection{Image-level Annotation}\label{sec:anno_tag}

Image-level annotation, or tag annotation, is a concise text or binary label assigned to each image, stands out as the most efficient form of annotation. 
Many image-level annotation-based methods draw inspiration from the concept of class activation mapping (CAM)~\cite{zhou2016learning}. 
Several approaches utilize CAM to generate object localization proposals or achieve pixel-wise segmentation of entire objects.
For \textbf{detection} tasks, Wang \textit{et al.}~\cite{hwang2016self} introduced a dual-branch network that concurrently optimizes classification and lesion detection. This approach involves supervising the CAM-based lesion detection network solely with image-level annotations. 
The two branches are guided in tandem through a weight-sharing technique, employing a weighting parameter to regulate the learning focus between classification and detection tasks. 
In the realm of lesion detection, Dubost \textit{et al.}~\cite{dubost2020weakly} proposed a weakly supervised regression network, validated on both 2D and 3D medical images. 
For the \textbf{segmentation} task, Li \textit{et al.}~\cite{li2022deep} proposed a breast tumor segmentation method with only image-level annotations based on CAM and deep-level set (CAM-DLS).
It integrates domain-specific anatomical information from breast ultrasound to reduce the search space for breast tumor segmentation.
Similarly, Li \textit{et al.}~\cite{li_sg-mian_2024} a self-guided multiple information aggregation network for skin lesion segmentation using multiple spatial perceptron solely using classification information as guidance to discriminate the key classification features of lesion areas.
Meanwhile, Chen \textit{et al.}~\cite{chen2022c} proposes a causal CAM method for organ segmentation, which is based on the idea of causal inference with a category-causality chain and an anatomy-causality chain.
In addition, several studies~\cite{lin2019seg4reg,lin2021seg4reg+} demonstrate that bridging the classification task and dense prediction task (e.g., detection and segmentation) via CAM-based methods is beneficial for both tasks.
Compared to natural images, medical images are usually from low contrast, limited texture, and varying acquisition protocols~\cite{zhang2021weakly}, which makes directly applying CAM-based methods less effective.
Fortunately, incorporating the clinical priors (e.g., objects' size~\cite{fruh2021weakly}) into the weakly supervised detection task is promising to improve the performance.
% ---------------------------------------------------------------
\subsubsection{Point Annotation}\label{sec:anno_point}
% ---------------------------------------------------------------
Point annotation refers to the annotation of a single point of an object.
Several studies~\cite{roth2021going,dorent2021inter,khan2019extreme} focus on using extreme points as the annotation to perform pixel-level segmentation.
These methods typically consist of three steps: 1) extreme point selection; 2) initial segmentation with a random walk algorithm; 3) training of the segmentation model with the initial segmentation results. 
The last two steps can be iterated until the segmentation results are stable.
However, these methods require the annotators to locate the boundary of the objects, which is still laborious in practice.
In contrast, other studies~\cite{yoo2019pseudoedgenet,zhao2020weakly,qu2020weakly,qu2019weakly,tian2020weakly,belharbi2021deep,lin2023nuclei,valvano2021learning} use center point annotation to perform pixel-level segmentation for the task of cell/nuclear segmentation. 
These methods typically adopt the Vorinor~\cite{kise1998segmentation} and cluster algorithms to perform coarse segmentation. 
Then different methods are used to refine the segmentation results, such as iterative optimization~\cite{qu2019weakly,qu2020weakly}, self-training~\cite{zhao2020weakly}, and co-training~\cite{lin2023nuclei}. 

Compared with full annotation, point annotation can reduce the annotation time by around 80\%~\cite{qu2020weakly}.
However, some issues have not been addressed. 
First, existing methods typically derived pseudo labels from the point annotation, which are based on strong constraints or assumptions (e.g., Voronoi) from the data, restricting the generalization of these methods to other datasets~\cite{lin2023nuclei}.
Second, due to the lack of explicit boundary supervision, there is a non-negligible performance gap between the weakly supervised methods with points and the fully supervised methods.

\subsubsection{Scribble Annotation}\label{sec:anno_scribble}
% ---------------------------------------------------------------
Scribble annotation, a set of scribbles drawn on an image by the annotators, has been recognized as a user-friendly alternative to bounding box annotation~\cite{tajbakhsh2020embracing}. 
Compared with point annotation, it provides the rough shape and size information of the objects, which is promising to improve the segmentation performance, especially for objects with complex shapes.
Wang \textit{et al.}~\cite{wang2018interactive} propose a self-training framework with differences in model predictions and user-provided scribbles. 
Can \textit{et al.}~\cite{can2018learning} develop a random walk algorithm that incrementally performs region growing method around the scribble ground truth, while
Lee \textit{et al.}~\cite{lee2019scribble2label} introduce Scribble2Label, a method that integrates the supervision signals from both scribble annotations and pseudo labels with the exponential moving average. 
Furthermore, Dorent \textit{et al.}~\cite{dorent2020scribble} extend the Scribble-Pixel method to the domain adaptation scenario, where a new formulation of domain adaptation is proposed based on CRF and co-segmentation with the scribble annotation. 
Zhang \textit{et al.}~\cite{zhang2022cyclemix} adopt mix augmentation and cycle consistency for the Scribble-Pixel method, demonstrating the improvement of both weakly and fully supervised segmentation methods.
Zhou \textit{et al.}~\cite{zhou_weakly_2023} proposed a scribble-supervised approach that combines superpixel-guided scribble walking with class-wise contrastive regularization to augment the structural priors into the weak annotations.

\subsubsection{Box Annotation}\label{sec:anno_box}

Box annotation encloses the segmented region within a rectangle, and various studies have focused on this Box-Pixel scenario. 
Rajchl \textit{et al.}~\cite{rajchl2016deepcut} employ a densely-connected random field (DCRF) with an iterative optimization method for MRI segmentation. 
Wang \textit{et al.}~\cite{wang2021accurate,wang2021bounding} adopt smooth maximum approximation based on the bounding box tightness prior~\cite{hsu2019weakly}, that is, an object instance should touch all four sides of its bounding box. 
Thus, a vertical or horizontal crossing line within a box yields a positive bag because it covers at least one foreground pixel. 
Studies~\cite{wang2021bounding} demonstrate that the Box-Pixel method yields promising performance, being only 1--2\% inferior to the fully supervised methods.
In a recent work, Wong \textit{et al.}~\cite{wong_scribbleprompt_2025} proposed ScribblePrompt, which is flexible to different annotation styles, including bounding boxes, points, and scribbles. Overall, box annotation offers a strong balance between efficiency and accuracy, but may struggle with irregular or overlapping shapes.

\subsection{Multi-instance Learning}

To address label granularity alignment from the task perspective, \textbf{multi-instance learning (MIL)} enables deep learning models to make fine-grained predictions using only coarse, image-level supervision. MIL organizes each image or specimen as a \textit{bag} of multiple \textit{instances} (e.g., patches, regions, or cells), with only the bag-level label observed during training. Under the standard MIL assumption, a bag is positive if at least one of its instances is positive. This framework supports weak supervision and allows for fine-grained analysis—such as localizing disease regions—even when only image-level labels are available.

Distinct from approaches that primarily calibrate the form of supervision, MIL is specifically tailored to bridge the granularity gap between global supervision and dense prediction requirements. By modeling the instance collections within each image, MIL enables inference of instance-level relevance from bag labels, thereby supporting fine-grained pattern recognition using only bag-level supervision.

In this survey, we categorize MIL approaches by their prediction target: \textbf{local-context MIL} focuses on detecting and classifying patterns at the instance level, while \textbf{global-context MIL} aims for bag-level predictions. Representative works are summarized in Appendix Tab.~\ref{tab:mil}.

\subsubsection{Local-Context MIL}
Within the MIL paradigm, local-context MIL approaches focus on the fine-grained identification and localization of specific disease patterns within medical images. These methods are designed to infer instance-level labels for individual patches, enabling precise delineation of pathological regions. By determining the status of each constituent patch, local-context MIL naturally encompasses global-context capabilities—the image-level classification emerges as a function of the detected local annotations. This hierarchical inference process allows clinicians not only to receive diagnostic outcomes but also to visualize the spatial distribution of disease manifestations, enhancing interpretability and clinical utility. The ability to simultaneously perform localization and diagnosis makes local-context MIL particularly valuable in applications requiring both detection sensitivity and anatomical precision.

Schwab \textit{et al.} \cite{schwab2020localization} apply MIL to chest X-ray classification and localization by processing image patches through a CNN to assess their probability of containing critical findings. While traditional MIL approaches integrate patch-level predictions using max-pooling or average-pooling, Couture \textit{et al.} \cite{couture2018multiple} enhance this by implementing a quantile function for pooling, providing better characterization of sample heterogeneity. The field has evolved toward learning-based aggregation methods, notably the attention-based MIL developed by Ilse \textit{et al.} \cite{ilse2018attention}, which better captures and interprets regions of interest. Wang \textit{et al.} \cite{wang2021learning} extend this concept with an inductive attention guidance network for pancreatic ductal adenocarcinoma, where the attention mechanism bridges global classification and local segmentation by identifying relevant regions.

Other intriguing improvements in local-context MIL are springing up as well. Researchers have tried many different ways to facilitate instance prediction \cite{dov2021weakly,manivannan2017subcategory,jia2017constrained,xu2019camel}. Dov \textit{et al.} \cite{dov2021weakly} address cytopathology challenges—where informative instances are sparse and exhibit varied abnormalities—by implementing maximum likelihood estimation to simultaneously predict bag-level labels, diagnostic scores, and instance-level labels. For retinal nerve fiber layer classification, Manivannan \textit{et al.} \cite{manivannan2017subcategory} overcome strong intra-class variation by mapping instances into a discriminative subspace that enhances feature disentanglement. Jia \textit{et al.} \cite{jia2017constrained} incorporate multi-scale image features to extract more latent information from histopathology images. Addressing the limitation of image-level-only labels in MIL, Xu \textit{et al.} \cite{xu2019camel} develop an automatic instance-level label generation method, creating a promising direction for local-context MIL algorithm development.

In parallel, there has been significant progress in related domains such as phenotype categorization \cite{yao2020whole, hashimoto2020multi, yao2019deep} and multi-label classification \cite{mercan2017multi}. These investigations have further exemplified the versatility and potential of the MIL algorithm in addressing complex challenges across various subfields. 

\subsubsection{Global-Context MIL}
In contrast to the localization focus of local-context approaches, global-context MIL aims to detect whether or not target patterns exist. For example, for the COVID-19 screening problem, researchers \cite{li2021novel} have designed MIL algorithms to classify an input sample as severe or not instead of locating every abnormal patch. 

To facilitate the prediction of image-level labels (\textit{e.g.} WSI-level label in the context of computational pathology), researchers normally start from one of two aspects, namely instance- and bag-level. Most existing MIL algorithms \cite{tomita2019attention, hashimoto2020multi,naik2020deep,lu2021data} are based on the basic assumption that instances of the same bag are independent and identically distributed. Consequently, the correlations among instances are neglected, which is not realistic. Several subsequent works have taken the correlation among instances or tissues into consideration \cite{shao2021transmil,wang2022lymph,wang2019rmdl,raju2020graph,han2020accurate}. In \cite{shao2021transmil}, Shao \textit{et al.} introduce Vision Transformer (ViT) into MIL for gigapixel WSIs due to its great advantage in capturing the long-distance information and correlation among instances in a sequence. Meanwhile, to conduct precise lymph node metastasis prediction, Wang \textit{et al.} \cite{wang2022lymph} not only incorporate a pruned Transformer into MIL but also develop a knowledge distillation mechanism based on other similar datasets, such as a papillary thyroid carcinoma dataset, effectively avoiding the overfitting problem caused by the insufficient number of samples in the original dataset. Similarly, Raju \textit{et al.} \cite{raju2020graph} design a graph attention MIL algorithm for colorectal cancer staging, which utilizes different tissues as nodes to construct graphs for instance relation learning. Further, in order to utilize the multi-resolution characteristics of WSIs, Shi \textit{et al.} \cite{shi2023structure} consider WSIs as multi-scale graphs and utilize attention mechanism to integrate their information for primary tumor stage prediction. Similar idea can be found in \cite{yan2023genemutation, shi2023mg, xiang2023multi}. Besides, Liu \textit{et al.} \cite{liu2024advmil} firstly propose an integration of GAN with MIL mechanism for robust and interpretable WSI survival analysis by more accurately estimating target distribution. 

For bag-level improvement, recent years have witnessed two feasible approaches, namely, improved pooling methods and pseudo bags. On the one hand, in order to aggregate the instances with the most information, some researchers have developed novel aggregation methods in MIL algorithms instead of the traditional max pooling \cite{chikontwe2020multiple,das2018multiple,jin2024hmil,guo2024focus}. For example, in \cite{chikontwe2020multiple}, the authors design a pyramid feature aggregation method to directly obtain a bag-level feature vector. On the other hand, however, there is an inherent problem for MIA, especially for histopathology --- the number of WSIs (bags) is usually small, while in contrast, one WSI has numerous patches, leading to an imbalance in the number of bags and instances. To address this problem, Zhang \textit{et al.} \cite{zhang2022dtfd} randomly split the instances of a bag into several smaller bags, called "pseudo bags", with labels that are consistent with the original bag. A similar idea can also be seen in \cite{li2021novel}. Moreover, Jin \textit{et al.} \cite{jin2024hmil} introduce a hierarchical multi-instance learning (HMIL) framework that enhances WSI classification through explicit modeling of the hierarchical relationships between instance-level and bag-level label distributions, creating a more cohesive cross-granularity learning paradigm.

Other improvements in MIL algorithms are also worth mentioning \cite{su2022attention2majority,tennakoon2019classification,wang2020ud}. In \cite{su2022attention2majority}, a novel sampling method is developed to collect instances with high confidence. This method excludes patches shared among different classes and tends to select the patches that match with the bag-level label. In \cite{tennakoon2019classification}, the authors utilize the extreme value theory to measure the maximum feature deviations and consequently leverage them to recognize the positive instances, while in \cite{wang2020ud}, Wang \textit{et al.} introduce an uncertainty evaluation mechanism into MIL for the first time, and train a robust classifier based on this mechanism to cope with OCT image classification problem. 

\subsection{Discussion}
Weakly supervised learning addresses the challenges of limited and inexact annotations in MIA by calibrating both the annotation form and granularity of supervision. Within this paradigm, annotation-efficient learning and MIL represent two complementary facets for coping with weak supervision.

Annotation-efficient learning leverages diverse annotation types—points, scribbles, boxes, and image tags—each suited to different object characteristics and annotation budgets. Points and scribbles are efficient for objects with uniform shape, while boxes handle morphological variation, and image tags offer the lowest annotation cost at the expense of spatial detail. Recent advances in HFMs, especially vision-language models, have enabled high-quality predictions from minimal supervision by integrating multi-modal cues and prompt-based learning~\cite{li2023blip,ren2024grounded}. Future directions include unifying multiple weak supervision signals, leveraging human-in-the-loop strategies, and mining knowledge from multi-modal data to further reduce annotation costs.

Multi-instance learning (MIL) addresses the calibration of annotation granularity, particularly for tasks such as gigapixel-sized WSI analysis, where only bag-level labels are available. MIL enables fine-grained, instance-level inference by modeling bag-instance and instance-instance relationships, effectively aligning global supervision with the need for precise localization or characterization. The rise of HFMs trained on large and diverse datasets has significantly enhanced MIL by providing rich, transferable features, boosting performance without extra annotation~\cite{vorontsov2024foundation,chen2024towards,lu2024visual,ma2024towards,xu2024multimodal,xu2024whole}. Key future directions include developing more interpretable and privacy-preserving MIL frameworks~\cite{javed2022additive,kapse2024si}, designing efficient attention mechanisms for high-dimensional features~\cite{guo2024histgen,li2024rethinking,tang2024feature,guo2024focus}, and integrating foundation models with domain-specific knowledge~\cite{zhang2023textadaptation,yin2024prompting,lu2024pathotune}.
\section{Label Refinement} \label{sec:al}

The \textit{label refinement} scenario arises when existing annotations are limited or suboptimal, often due to annotation costs and expert scarcity in MIA. \textbf{Active learning (AL)} has emerged as a primary solution by prioritizing expert annotation on the most informative samples. Unlike prior paradigms that passively exploit available labels, AL actively involves human experts to iteratively improve data quality. Typically, AL starts with a model trained on a small, initially labeled dataset and, in each cycle, ranks the remaining unlabeled or weakly labeled samples by informativeness for expert review. The model is then retrained with the refined labels, and this process repeats until the desired performance or annotation budget is reached.

Informative sampling is central to AL for label refinement, as its effectiveness relies on selecting samples whose refined labels most benefit the model. Accordingly, AL methods are typically classified by their \textbf{informativeness evaluation} criteria, such as uncertainty- or representativeness-based, and \textbf{sampling strategies} for efficient querying. Appendix Table~\ref{tab:al} summarizes representative methods. The following subsections detail these criteria and strategies.

\begin{figure}[htbp]
	\centering
\includegraphics[width=\textwidth]{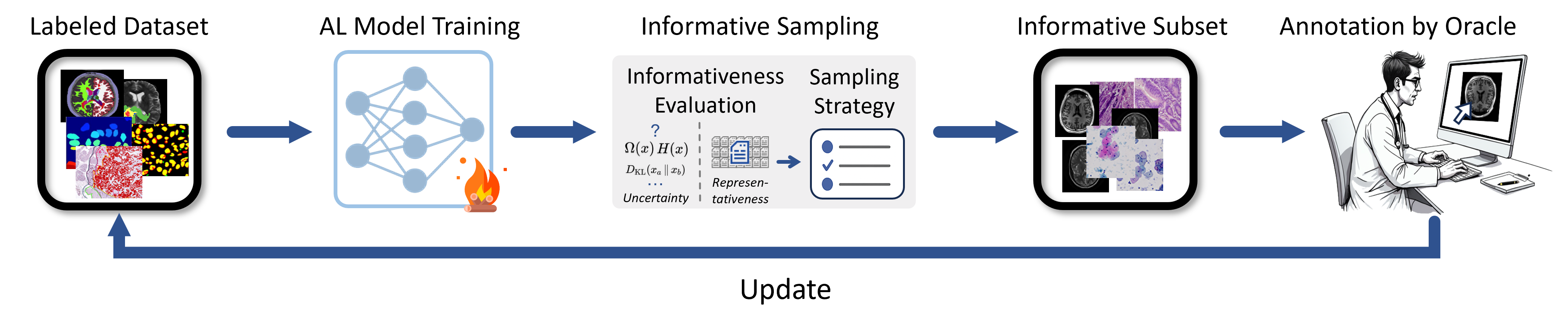}
	\caption{Overview of the active learning (AL) paradigm, which addresses the \textit{label refinement} scenario. AL iteratively trains a model by employing informative sampling strategies to select and annotate the most valuable unlabeled samples, thereby maximizing model performance under a limited annotation budget.}
\label{fig_al_schematic}
\end{figure}

\subsection{Informativeness Evaluation}

Informativeness evaluation is fundamental to active learning, guiding the selection of unlabeled samples that are most beneficial for model improvement. Existing methods primarily fall into two categories: uncertainty-based and representativeness-based strategies.

Uncertainty-based methods select samples where the model is least confident, employing measures such as entropy, margin, or Bayesian uncertainty~\cite{yang2017suggestive, chen2016dcan}. These techniques have evolved to incorporate feature-space similarity and loss prediction networks, especially in structured tasks like segmentation, where metrics such as the Dice coefficient are used to estimate sample informativeness~\cite{zhao2021dsal, wu2021covid}. Recent advances further enhance robustness by integrating gradient-based uncertainty and out-of-distribution detection~\cite{hu2023learning, schmidt2024focused}.

Representativeness-based approaches aim to ensure that selected samples capture the diversity of the data distribution. Clustering, curriculum learning, and latent space distance metrics have been widely adopted for this purpose~\cite{li2021pathal, guo2018curriculumnet}. In medical imaging, where class imbalance and label noise are common, mechanisms such as automatic exclusion of noisy samples and task-specific loss functions are often incorporated~\cite{huang2019o2u, liu2017stein}.

A growing body of work seeks to unify these perspectives through hybrid criteria that combine uncertainty, representativeness, and other informativeness signals~\cite{qu2023openal, mahapatra2024gandalf, li2023hal}. Such methods leverage, for example, Mahalanobis distance, graph-based propagation, or divergence-based measures to improve sample selection efficiency and model robustness ~\cite{tang2023pld, atzeni2022deep}. In summary, informativeness evaluation in AL has thus evolved from simple uncertainty heuristics to hybrid strategies that jointly consider uncertainty, representativeness, and deep feature representations, significantly improving annotation efficiency and model robustness.

\subsection{Sampling Strategy}
Once informativeness has been assessed, the next challenge lies in designing effective sampling strategies that maximize annotation efficiency, particularly when labeling resources are limited.

A straightforward approach is top-$k$ selection, where samples with the highest informativeness scores are prioritized. However, this can lead to redundancy, as similar samples may be repeatedly chosen~\cite{zhou2021active}. To overcome this, many later strategies integrate additional criteria such as diversity and representativeness into the selection process~\cite{ozdemir2021active, zhao2017infovae}. By jointly considering these factors, hybrid methods aim to ensure that selected samples are both informative and broadly representative of the underlying data distribution—a consideration especially important in domains characterized by data heterogeneity or class imbalance.

To further enhance the sampling process, some approaches incorporate data augmentation or generative models, increasing the diversity of candidate samples and reducing annotation redundancy~\cite{mahapatra2018efficient}. Optimization-based formulations have also been proposed, dynamically balancing uncertainty, consistency, and diversity during selection~\cite{li2023hal, qu2023openal}. Other strategies leverage feature-space metrics, clustering, or graph-based algorithms to maximize coverage and minimize overlap among selected samples~\cite{bai2023slpt, mahapatra2024gandalf, mahapatra2024alfredo}.

Overall, effective sampling strategies in active learning seek to maximize annotation value by balancing informativeness, diversity, and representativeness~\cite{schmidt2024focused, wang2024dual, tang2023pld}. These principles have been shown to improve label efficiency and model robustness, particularly in complex and data-rich applications.

\subsection{Discussion}
Active learning addresses label scarcity in medical imaging by iteratively selecting the most informative samples for expert annotation, a process shaped by the interplay between informativeness evaluation and sampling strategy. Early in the annotation process, model performance is largely determined by how well informativeness is quantified, with advanced criteria offering clear gains over naive uncertainty measures~\cite{wang2024comprehensive}. As the labeled set expands, redundancy emerges as the main bottleneck, making sophisticated sampling strategies increasingly important \cite{follmer2024active}.

The rise of HFMs, particularly multi-modal architectures such as MedSAM \cite{ma2024segment} is fundamentally reshaping AL by providing rich, transferable representations that enhance both informativeness evaluation and sampling efficiency~\cite{gupte2024revisiting}. HFMs enable dynamic adjustment of an image’s uncertainty profile through textual or visual prompts, allowing rare or long-tail cases to be surfaced that traditional approaches may overlook. However, confidence estimates from HFMs often require careful calibration, especially for atypical cases ~\cite{gupte2024revisiting}. Future frameworks should therefore jointly optimize prompt design and sample selection with careful calibration of model confidence, leveraging the capabilities of HFMs to achieve more efficient and clinically relevant annotation.

\section{Challenges and Future Directions} \label{sec:cnfd}
\begin{figure}[htbp]
\centering
\includegraphics[width=\textwidth]{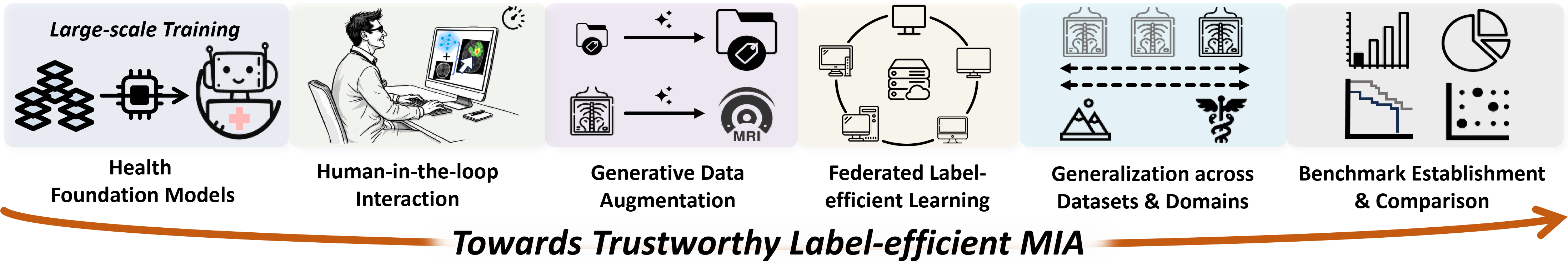} 
\caption{Future directions for label-efficient MIA are mapped along the research-to-deployment continuum, encompassing the development of health foundation models, human-in-the-loop interaction, generative data augmentation, federated label-efficient learning, cross-domain generalization, and the establishment of standardized benchmarks for clinical evaluation.}
	\label{fig_future}
\end{figure}
While the previous sections have reviewed various label-efficient learning paradigms designed for MIA, notable challenges still impede their translation from research to clinical practice. Addressing these barriers is vital for realizing the full potential of these methods in real-world healthcare settings. In the following, we summarize the main challenges and discuss promising future directions.

\subsection{Health Foundation Models}
The diversity and complexity of MIA tasks, coupled with persistent labeled data scarcity, have historically limited the scalability and clinical translation of label-efficient algorithms. Recent advances in HFMs have shifted this paradigm by introducing highly generalizable representations pretrained on heterogeneous medical image datasets~\cite{he2024foundation, chen2024towards, wu2024voco, xiang2025vision}. This extensive and diverse pretraining enables HFMs to capture rich, transferable visual features, thereby substantially reducing the volume of annotated data and expert time required for downstream applications. As a result, HFMs facilitate rapid adaptation to tasks such as classification, segmentation, detection, and report generation, often achieving strong performance with minimal supervision~\cite{ma2024segment, zhang2025multimodal}. Collectively, these developments position HFMs as a foundational technology in the field, delivering significant improvements in annotation efficiency, generalizability, and clinical applicability for label-efficient MIA. Despite these advances, however, the reliance of HFMs on downstream fine-tuning with task-specific labeled data remains a fundamental limitation, as truly label-efficient or label-free learning has yet to be fully realized. These technical limitations are further compounded by broader issues, such as the scalable and ethical curation of large-scale medical datasets, the mitigation of biases arising from data heterogeneity, and the integration of cross-modal and longitudinal data sources—such as genomics profile, electronic health records, and pathology whole-slide images. Furthermore, ensuring that HFMs are aligned with clinical safety standards and are sufficiently interpretable is essential for their reliable deployment in real-world healthcare settings. Addressing these challenges is critical for HFMs to progress from large-scale models to foundational infrastructure supporting the future of label-efficient MIA.

\subsection{Human-in-the-loop Interaction}

The introduction of HFMs pretrained on large-scale medical data has facilitated more efficient human-in-the-loop (HITL) collaboration in medical image analysis, particularly for reducing expert annotation costs. While AL focuses on selecting informative samples for annotation during training, HITL strategies such as interactive segmentation incorporate real-time expert feedback at inference. Beyond technical benefits, recent studies highlight that effective HITL in medical AI requires supporting meaningful human control and co-reasoning, integrating ethical and practical judgment from both clinicians and patients into the decision-making process \cite{salloch2024humans, marinov2024deep}. Meanwhile, in the broader machine learning community, there has been substantial progress in label-efficient HITL paradigms, including adaptive querying, scalable annotation platforms, and uncertainty-aware feedback mechanisms, with recent frameworks demonstrating that verification of auto-suggested labels can be 3–4 times faster than manual labeling and improve overall efficiency by 1.5–2 times \cite{beck2024beyond, feng2025duo}. However, these advances are still underexplored in MIA, where annotation costs, privacy, and workflow integration pose unique challenges. Bridging these advances, future research should focus on adapting state-of-the-art HITL strategies to the medical domain by leveraging foundation models for automated pre-annotation, developing intuitive and privacy-preserving expert collaboration interfaces, and embedding interactive feedback into clinical workflows \cite{huang2025pathologist}. Such efforts can further reduce annotation effort, enhance model robustness, and accelerate the safe and effective deployment of AI in clinical settings.

\subsection{Generative Data Augmentation}
With robust annotation pipelines established through human-in-the-loop interaction, the next step is to address the long-tail of medical imaging datasets by synthesizing rare or underrepresented cases. Synthetic data generated by advanced models has proven effective for enhancing diversity and improving performance in vision tasks \cite{goceri2023medical}. Although early GAN-based methods suffer from instability and limited diversity \cite{lin2022insmix}, current research increasingly adopts two main architectures that offer higher fidelity, broader mode coverage, and more stable optimization. Diffusion models \cite{ho2020denoising} have demonstrated clear benefits in label‑efficient multi‑organ CT segmentation \cite{wang2024textguideddiff}, weakly‑supervised medical image segmentation \cite{xu2023conditional} and brain MRI label rectification \cite{li2024diffrect}, where they deliver 2-5\% gains in Dice or accuracy under limited annotation budgets. Meanwhile, MaskGIT-style transformers enable rapid, class-conditioned synthesis and achieve strong results with far less labeled data \cite{liu2024maskgitmed}. Despite these advances, several key challenges remain. First, it is critical to devise principled strategies for balancing real and synthetic samples to prevent overfitting to model-generated artifacts. Second, scaling generative augmentation to high-resolution modalities such as WSI presents significant computational and architectural hurdles. Third, integrating expert-driven, text- or report-level conditioning remains an open problem for enabling controllable and clinically meaningful data synthesis. Addressing these challenges will be essential for establishing generative augmentation as a robust and foundational component of future label-efficient MIA pipelines.
%The expressive priors encoded by HFMs facilitate high-fidelity generative data augmentation. By leveraging pretrained decoders, these models can synthesize anatomically coherent and clinically plausible rare cases, thereby enhancing model robustness in the face of data imbalance and sparsity.

\subsection{Federated Label-efficient Learning} 

As the scale of medical imaging data continues to grow, privacy concerns and the need for cross-institutional collaboration rapidly emerge as critical bottlenecks. \textbf{Federated learning (FL)} provides an effective framework to address these challenges by enabling collaborative model training across institutions without sharing raw data, thereby safeguarding patient privacy. More importantly, FL facilitates label-efficient learning by allowing institutions to maximize the utility of their limited labeled data in a distributed setting \cite{rieke2020future}. This approach has yielded fruitful results in The field of MIA \cite{li2020multi,dayan2021federated,lu2022federated}. However, current FL algorithms are primarily trained in a supervised manner. When applying the FL to real-world scenarios in MIA, a crucial problem, namely, label deficiency, may appear in local health datasets. Labels may be missing to varying degrees between medical centers, or the granularity of the labels will vary. Therefore, designing label-efficient federated learning methods to address this significant problem is an important research direction. For example, incorporating informative sampling strategies from active learning \cite{chen2024think, wu2024feda3i} into FL frameworks shows great potential for improving label efficiency in this setting. Nonetheless, FL faces unique challenges such as communication delays and efficiency; while recent works explore communication-efficient strategies \cite{yan2025fedvck}, future research should investigate approaches such as adapters or low-rank updates, which can be communicated across institutions to jointly optimize both label efficiency and communication efficiency.

\subsection{Generalization Across Domains and Datasets}
While recent advances in label-efficient MIA have reduced annotation requirements, the central challenge remains ensuring robust and reliable clinical deployment across diverse environments. Even HFMs can suffer significant performance drops when deployed on previously unseen scanners, imaging protocols, or patient populations \cite{he2024foundation}. This vulnerability is largely due to hardware heterogeneity, acquisition variability, and demographic differences, which constrain the transferability of models trained on limited or homogeneous datasets \cite{yoon2024domain}. To address this, \textbf{domain generalization (DG)} must become a central design objective for future label-efficient MIA systems. Recent research focuses on systematically identifying and mitigating domain biases that persist after large-scale pre-training, with growing emphasis on adaptive model recalibration ~\cite{zu2024embedded, chen2024each, jiang2025generalizable}. These advances collectively aim to ensure that models retain reliability and generalizability under label efficiency, which is essential for clinically robust and globally deployable MIA solutions.

\subsection{Benchmark Establishment and Comparison}
As the preceding sections have systematically reviewed the main challenges and promising directions in label-efficient MIA, their practical value, however, remains difficult to assess due to the lack of standardized and rigorous evaluation. The current landscape is highly fragmented, with studies differing in tasks, target organs, annotation budgets, protocols, and data partitioning \cite{bassi2024touchstone}. Privacy constraints further restrict access to large and diverse datasets, limiting evaluation to single-center cohorts with non-standardized splits \cite{jin2024fairmedfm}.  As a result, published results seldom reveal which method achieves the best accuracy per annotation cost in realistic clinical scenarios. Overcoming these limitations requires a new benchmarking infrastructure with curated multi-task datasets, fixed label budgets, cost-aware metrics for annotation and computational efficiency, and standardized protocols for human-in-the-loop evaluation. Federated benchmarking, where the code runs securely within each institution, can enable privacy-preserving comparison among multiple centers. Continuous, blinded leaderboards are also crucial for reproducibility and bias detection. Establishing a unified, community-driven benchmarking framework with these features is essential for objective model comparison and for translating methodological advances into reliable, clinically deployable systems.

\section{Conclusion}\label{sec:con}
Label-efficient learning has emerged as a pivotal direction in medical image analysis, addressing not only the practical constraints of annotation scarcity but also prompting a re-examination of the fundamental relationship between data, supervision, and clinical value. In this survey, we have introduced a unified taxonomy that spans scenarios of no label, insufficient label, inexact label, and label refinement. This framework has clarified the methodological landscape and illuminated the distinct challenges and opportunities inherent to each paradigm. Through a critical analysis of state-of-the-art methodologies, we have highlighted both recent advances and the persistent barriers to clinical adoption. Our synthesis demonstrates that genuine progress in label efficiency requires not only algorithmic innovation but also coordinated advances in large-scale data curation, adaptive learning strategies, and standardized evaluation. As the field moves toward broader clinical integration, the challenges of generalization, interoperability, and meaningful assessment remain central. By clarifying these issues and outlining future directions, this survey aims to serve as both a reference and a foundation for the continued evolution of label-efficient learning in medical imaging.

%%
%% The next two lines define the bibliography style to be used, and
%% the bibliography file.
\bibliographystyle{ACM-Reference-Format}
\bibliography{reference}

%%
%% If your work has an appendix, this is the place to put it.
\appendix
\newpage
\noindent\Large{\textbf{Appendix}}
\section{Survey Scope}
\label{appendix1}
To ensure comprehensive coverage of relevant literature, we conducted a systematic search using Google Scholar for publications related to label-efficient medical imaging up to March 2025. Additionally, we queried the arXiv preprint server for manuscripts containing key terms pertinent to label-efficient learning in medical imaging. Major conference proceedings—including CVPR, ICCV, ECCV, NeurIPS, AAAI, and MICCAI—as well as leading journals such as Medical Image Analysis (MIA), IEEE Transactions on Medical Imaging (TMI), and Nature Biomedical Engineering, were thoroughly reviewed based on paper titles and abstracts. Furthermore, the reference lists of selected papers were examined to identify additional relevant works. In cases where similar work appeared in multiple venues, only the most impactful or comprehensive publication(s) were included in this survey.

\clearpage

\section{Surveyed Literature and Datasets}

\subsection{Surveyed Literature}
\begin{table*}[ht]
\centering
\caption{Surveyed Self-supervised Learning-based Studies in Medical Image Analysis.}

\resizebox{\textwidth}{.56\textwidth}{
\begin{threeparttable}
\begin{tabular}{@{}llllll@{}}
\toprule
 &Reference & Organ & Proxy Task Design & Dataset & Publication \\ \midrule
\rule{0pt}{2ex} \multirow{20}{*}{\rotatebox{90}{Classification}}&  Li \textit{et al.} \cite{li2020self} & Retina & Multi-modal Contrastive Learning  & ADAM; PALM& TMI 2020\\
\cline{2-6}
% \rule{0pt}{2.5ex}&Zhao \textit{et al.} \cite{zhao2021anomaly}  & Retina; Lung & Inpainting; Local Pixel Shuffling;  & RetinalOCT; ChestX &RetinalOCT: AUC: 0.9642; F1: 0.9342\\
% &&&Non-Linear Intensity Transformation&&ChestX: AUC: 0.8265; F1: 0.8214\rule[-1.2ex]{0pt}{0pt}\\
% \cline{2-6}
\rule{0pt}{2.5ex}&Koohbanani \textit{et al.} \cite{koohbanani2021self} & Breast;  &Magnification Prediction;  & CAMELYON 2016; &TMI 2021  \\
&&Cervix;&Solving Magnification Puzzle;&KATHER;& \\
&&Colon&Hematoxylin Channel Prediction&Private Dataset: 217 Images&\rule[-1.2ex]{0pt}{0pt}\\
\cline{2-6}
% \rule{0pt}{2.5ex}&Li \textit{et al.} \cite{li2021rotation} & Retina &Image Rotation  & ADAM; PALM; DRD & ADAM: AUC: 0.7811; PALM: AUC: 0.9912 \rule[-1.2ex]{0pt}{0pt}\\
% \cline{2-6}
\rule{0pt}{2.5ex}&Azizi \textit{et al.} \cite{azizi2021big} & Skin; Lung &Multi-Instance Contrastive Learning & Priavte Dermatology Dataset; CheXpert  &CVPR 2021\rule[-1.2ex]{0pt}{0pt}\\
\cline{2-6}
% \rule{0pt}{2.5ex}&Yang \textit{et al.} \cite{yang2022cs} & Colon & Cross-stain prediction + Contrastive Learning& KATHER& Acc: 0.918\rule[-1.2ex]{0pt}{0pt}\\
% \cline{2-6}
\rule{0pt}{2.5ex}&Tiu \textit{et al.} \cite{tiu2022expert} & Lung & Contrastive Learning  & CheXpert&Nature BME 2022\rule[-1.2ex]{0pt}{0pt}\\
\cline{2-6}
\rule{0pt}{2.5ex}&Chen \textit{et al.} \cite{chen2022scaling} & Breast; Lung; Kidney & Contrastive Learning  & TCGA-BRCA; TCGA-NSCLS; TCGA-RCC&CVPR 2022\rule[-1.2ex]{0pt}{0pt}\\
\cline{2-6}
\rule{0pt}{2.5ex}&Mahapatra \textit{et al.} \cite{mahapatra2022self} & Lymph; Lung; & Contrastive Learning Variant  & CAMELYON 2017; DRD; GGC&TMI 2022\\
&&Retina; Prostate&&\rule[-1.2ex]{0pt}{0pt}\\
\cline{2-6}
\rule{0pt}{2.5ex}&Wang \textit{et al.} \cite{wang2023ssd} & Skin & Self-supervised Knowledge Distillation  & ISIC 2019 &MIA 2023 \rule[-1.2ex]{0pt}{0pt}\\
\cline{2-6}
\rule{0pt}{2.5ex}& Huang \textit{et al.} \cite{huang2024systematic}  &  Multi-Organ   & SimCLR, MOCOv2,
SwAV, BYOL, SimSiam, DINO, BarlowTw & TissueMNIST; PathMNIST; TMED-2; AIROGS& CVPR 2024 \rule[-1.2ex]{0pt}{0pt}\\
\cline{2-6}
\rule{0pt}{2.5ex}& Tang \textit{et al.} \cite{tang2024self}  &  Lung & Self-Supervised Representation Distribution Learning & TCGA-EGFR; TCGA-Lung; Private Lung Dataset & TMI 2024 \rule[-1.2ex]{0pt}{0pt}\\
\cline{2-6}
\rule{0pt}{2.5ex}& Vorontsov \textit{et al.} \cite{vorontsov2024foundation}  & Multi-Organ & Self-distillation + Masked Image Modeling (DINOv2) & Cancer-related Diagnosis Datasets & Nat. Med. 2024 \rule[-1.2ex]{0pt}{0pt}\\
\cline{2-6}
\rule{0pt}{2.5ex}& Chen \textit{et al.} \cite{chen2024towards}  & Multi-Organ & Self-distillation + Masked Image Modeling (DINOv2) & Cancer-related Diagnosis Datasets & Nat. Med. 2024 \rule[-1.2ex]{0pt}{0pt}\\
\cline{2-6}
\rule{0pt}{2.5ex}& Lu \textit{et al.} \cite{lu2024visual}  & Multi-Organ & Visual-language Contrastive Learning + Captioning (CoCa)  & Cancer-related Diagnosis Datasets & Nat. Med. 2024 \rule[-1.2ex]{0pt}{0pt}\\
\hline
\rule{0pt}{2.5ex} \multirow{14}{*}{\rotatebox{90}{Segmentation}}& Hervella \textit{et al.} \cite{hervella2018retinal}$_{2018}$ & Retina &Multi-modal Reconstruction & Isfahan MISP & MICCAI 2018 \rule[-1.2ex]{0pt}{0pt}\\
\cline{2-6}
\rule{0pt}{2.5ex}&Spitzer \textit{et al.} \cite{spitzer2018improving}$_{2018}$ & Brain & Patch Distance Prediction   & BigBrain & MICCAI2018\rule[-1.2ex]{0pt}{0pt}\\
\cline{2-6}
\rule{0pt}{2.5ex} &  Bai \textit{et al.} \cite{bai2019self} & Heart & Anatomical Position Prediction & Private Dataset: 3825 Subjects & MICCAI 2019\rule[-1.2ex]{0pt}{0pt}\\
\cline{2-6}
\rule{0pt}{2.5ex}& Sahasrabudhe \textit{et al.} \cite{sahasrabudhe2020self}  & Multi-Organ & WSI Patch Magnification Identification &MoNuSeg & MICCAI 2020\rule[-1.2ex]{0pt}{0pt}\\
\cline{2-6}
\rule{0pt}{2.5ex}& Tao \textit{et al.} \cite{tao2020revisiting} & Pancreas &  Rubik's Cube Recovery& NIH PCT; MRBrainS18& MICCAI 2020\rule[-1.2ex]{0pt}{0pt}\\
\cline{2-6}
\rule{0pt}{2.5ex}&Lu \textit{et al.} \cite{lu2021volumetric}& Brain & Fiber Streamlines Density Map Prediction;& dHCP &MIA 2021\\
&&& Registration-based Segmentation Imitation& & \rule[-1.2ex]{0pt}{0pt}\\
\cline{2-6}
\rule{0pt}{2.5ex}&Tang \textit{et al.} \cite{tang2022self} & Abdomen; Liver; &Contrastive Learning; Masked Volume Inpainting;    & DECATHLON; &CVPR 2022\\
&&Prostate&3D Rotation Prediction &BTCV&\rule[-1.2ex]{0pt}{0pt}\\
\cline{2-6}
\rule{0pt}{2.5ex}&Jiang \textit{et al.} \cite{jiang2023anatomical} & Multi-organ &Anatomical-invariant Contrastive Learning    & FLARE 2022; BTCV &CVPR 2023\rule[-1.2ex]{0pt}{0pt}\\
\cline{2-6}
\rule{0pt}{2.5ex}&He \textit{et al.} \cite{he2023geometric} & Heart; Artery; Brain &Geometric Visual Similarity Learning    & MM-WHS-CT; ASOCA; CANDI; STOIC &CVPR 2023\rule[-1.2ex]{0pt}{0pt}\\
\cline{2-6}
\rule{0pt}{2.5ex}&Liu \textit{et al.} \cite{liu2023hierarchical} & Tooth &Hierarchical Global-local Contrastive Learning    & Private Dataset: 13,000 Scans &TMI 2023\rule[-1.2ex]{0pt}{0pt}\\
\cline{2-6}
\rule{0pt}{2.5ex}&Zheng \textit{et al.} \cite{zheng2023msvrl} & Multi-Organ &Multi-scale Visual Representation Self-supervised Learning    & BCV; MSD; KiTS &TMI 2023\rule[-1.2ex]{0pt}{0pt}\\
\cline{2-6}
\rule{0pt}{2.5ex}&Peng \textit{et al.} \cite{peng2024boundary} & Heart; Prostate & Contrastive Learning  & ACDC; PROMISE12 & MIA 2024\rule[-1.2ex]{0pt}{0pt}\\
\cline{2-6}
\rule{0pt}{2.5ex}&Purma \textit{et al.} \cite{purma2024genselfdiff} & Multi-Organ & Diffusion-based Reconstruction & Head and Neck Cancer; GlaS; MoNuSeg & TMI 2024\rule[-1.2ex]{0pt}{0pt}\\
\hline
\rule{0pt}{2.5ex} \multirow{6}{*}{\rotatebox{90}{Regression}} & Abbet \textit{et al.} \cite{abbet2020divide}  & Gland &Image Colorization & Private Dataset: 660 Images&MICCAI 2020 \rule[-1.2ex]{0pt}{0pt}\\
\cline{2-6}
\rule{0pt}{2.5ex}& \multirow{3}{*}{Srinidhi \textit{et al.} \cite{srinidhi2022self}}  & Breast; & WSI Patch Resolution Sequence & BreastPathQ; &MIA 2022\\
&& Colon& Prediction&CAMELYON 2016; &  \\
&&&&KATHER &  \rule[-2ex]{0pt}{0pt}\\
\cline{2-6}
\rule{0pt}{2.5ex}&Fan \textit{et al.} \cite{fan2023cancerself}  &  Brain; Lung  & Image Colorization; Cross-channel   &  GBM; TCGA-LUSC; NLST &TMI 2023\rule[-1.2ex]{0pt}{0pt}\\
\hline
\rule{0pt}{2.5ex} \multirow{32}{*}{\rotatebox{90}{Others}} & Zhuang \textit{et al.} \cite{zhuang2019selfsupervised} & Brain &  Rubik's Cube Recovery & BraTS 2018; Private Dataset: 1,486 Images & MICCAI 2019\rule[-1.2ex]{0pt}{0pt}\\ 
\cline{2-6}
\rule{0pt}{2.5ex}& \multirow{3}{*}{Chen \textit{et al.} \cite{chen2019self}}  & \multirow{3}{*}{Multi-Organ} & \multirow{3}{*}{Disturbed Image Context Restoration} & Private Fetus Dataset: 2,694 Images; &MIA 2019\\
&&&&Private Multi-organ Dataset: 150 Images;& \\
&&&&BraTS 2017& \rule[-1.2ex]{0pt}{0pt}\\
\cline{2-6}
\rule{0pt}{2.5ex}&Zhao \textit{et al.} \cite{zhao2020smore}  &  Brain  & Super-resolution Reconstruction   &  Private Dataset: 47 Images &TMI 2020\rule[-1.2ex]{0pt}{0pt}\\
\cline{2-6}
\rule{0pt}{2.5ex}&Li \textit{et al.} \cite{li2020sacnn} & Abdomen  & CT Reconstruction  & LDCTGC &TMI 2020\rule[-1.2ex]{0pt}{0pt}\\
\cline{2-6}
\rule{0pt}{2.5ex}&Cao \textit{et al.} \cite{cao2020auto}  & Brain & Missing Modality Synthesis & BraTS 2015; ADNI & AAAI 2020\rule[-1.2ex]{0pt}{0pt}\\ 
\cline{2-6}
\rule{0pt}{2.5ex}&Haghighi \textit{et al.} \cite{haghighi2020learning} & Lung &  Self-Discovery + Self-Classification  & LUNA; LiTS; CAD-PE; BraTS 2018; &   MICCAI 2020   \\
&&&+Self-Restoration& ChestX-ray14; LIDC-IDRI; SIIM-ACR& \rule[-1.2ex]{0pt}{0pt}\\
\cline{2-6}
\rule{0pt}{2.5ex}&Taleb \textit{et al.} \cite{taleb20203d}& Brain; Retina;  & 3D Contrastive Predictive Coding; 3D Jigsaw Puzzles; & BraTS 2018; & NeurIPS 2020\\
&&Pancreas &3D Rotation Prediction; 3D Exemplar Networks&DECATHLON;& \\
&&&Relative 3D Patch Location;&DRD& \rule[-1ex]{0pt}{0pt}\\
\cline{2-6}
\rule{0pt}{2.5ex}&Li \textit{et al.} \cite{li2021single} &  Breast; Pancreas; Kidney &Super-resolution Reconstruction;  Color Normalization  & WTS; Private Dataset: 533 Images& MIA 2021 \rule[-1.2ex]{0pt}{0pt}\\
\cline{2-6}
\rule{0pt}{2.5ex}&Wang \textit{et al.} \cite{wang2021transpath} & Multi-Organ & Contrastive Learning & TCGA; KATHER; MHIST & MICCAI 2021\\
&&&&PAIP; PatchCAMELYON& \rule[-1.2ex]{0pt}{0pt} \\
\cline{2-6}
\rule{0pt}{2.5ex}&Zhou \textit{et al.} \cite{zhou2021preservational}& Lung; Brain; Liver & Contrastive Learning + Image Reconstruction & ChestX-ray14; CheXpert; LUNA &  CVPR 2021\\
&&&&BraTS 2018; LiTS;& \rule[-1.2ex]{0pt}{0pt}\\
\cline{2-6}
\rule{0pt}{2.5ex}&Yan \textit{et al.} \cite{yan2022sam}& Multi-Organ  & Global and Local Contrastive Learning  & DeepLesion; NIH LN; Private Dataset: 94 Patients& TMI 2022\rule[-1.2ex]{0pt}{0pt}\\
\cline{2-6}
\rule{0pt}{2.5ex}&Haghighi \textit{et al.} \cite{haghighi2022dira}& Lung & Contrastive Learning + Reconstruction + & ChestX-ray14; CheXpert; & CVPR 2022 \\
&&&Adversarial Learning & Montgomery&  \rule[-1.2ex]{0pt}{0pt}\\
\cline{2-6}
\rule{0pt}{2.5ex}&Cai \textit{et al.} \cite{cai2023dualself}& Lung; Brain; Retina& Dual-Distribution Reconstruction & RSNA-Lung; LAG; VinDr-CXR; Brain Tumor MRI;& MIA 2023 \\
&& &&Private Lung Dataset: 5,000 Images& \rule[-1.2ex]{0pt}{0pt}\\
\cline{2-6}
\rule{0pt}{2.5ex}&Li \textit{et al.} \cite{li2023generic}& Retina  & Frequency-boosted Image Enhancement  &EyePACS;& MIA 2023 \\
&&&&Private Dataset: more than 10,000 Images& \rule[-1.2ex]{0pt}{0pt}\\
% \cline{2-6}
% \rule{0pt}{2.5ex}&Xie \textit{et al.} \cite{xie2022unimiss}$_{2022\text{CS}}$ & Multi-Organ & Contrastive Learning  & BCV; RICORD; &BCV: DSC: 0.8499; RICORD: AUC: 0.8906;\\
% &&&&JSRT Database; ChestXR&JSRT Database: DSC: 0.9408; ChestXR: AUC: 0.9907\rule[-1.2ex]{0pt}{0pt}\\
\bottomrule
\end{tabular}
\begin{tablenotes}    
        \footnotesize               
        \item[1] For the sake of brevity, we denote references that contain more than one task in the following abbreviations: \textbf{C}: Classification, \textbf{S}:Segmentation, \textbf{D}:Detection, \textbf{SR}: Super-resolution, \textbf{DN}: Denoising, \textbf{IT}: Image Translation, \textbf{RE}: Registration. 
      \end{tablenotes}
\end{threeparttable}
}
\label{tab:self}
\end{table*}
\begin{table*}[ht]
\centering
\caption{Overview of Semi-supervised Learning-based Studies in Medical Image Analysis.}

\resizebox{\textwidth}{.49\textwidth}{
\begin{threeparttable}
\begin{tabular}{@{}llllll@{}}
\toprule
 &Reference & Organ &  Semi-SL Algorithm Design & Dataset & Publication \\ \midrule
\rule{0pt}{2ex} \multirow{20}{*}{\rotatebox{90}{Classification}}&  Madani \textit{et al.} \cite{madani2018semi}$_{2018}$ & Lung & Semi-superivsed GAN & NIH PLCO; NIH Chest X-Ray&  ISBI 2018\rule[-1.2ex]{0pt}{0pt} \\ 
\cline{2-6}
\rule{0pt}{2.5ex}&  \rule{0pt}{2.5ex}Diaz-Pinto \textit{et al.} \cite{diaz2019retinal} &  Retina  & Semi-supervised DCGAN  & ORIGA-light; DRISHTI-GS; RIM-ONE; HRF; DRD; &TMI 2019\\
&&&&sjchoi86-HRF; ACRIMA; DRIVE; Messidor &\rule[-1.2ex]{0pt}{0pt}\\
\cline{2-6}
% \rule{0pt}{2.5ex}& Gyawali \textit{et al.} \cite{gyawali2019semi} & Lung & Temporal Ensembling + VAE  & Chexpert& AUC: 0.6697\rule[-1.2ex]{0pt}{0pt}\\
% \cline{2-6}
% \rule{0pt}{2.5ex}& Su \textit{et al.} \cite{su2019local} & Nuclei & Mean Teacher + Label Propagation& MoNuseg; Private Dataset: 17516 nuleus&MoNuseg: F1: 0.783; Private Dataset: F1: 0.7991\rule[-1.2ex]{0pt}{0pt}\\
% \cline{2-6}
\rule{0pt}{2.5ex}& Shi \textit{et al.} \cite{shi2020graph} & Lung; Breast & Graph Temporal Ensembling  & TCGA-Lung; TCGA-Breast&MIA 2020\rule[-1.2ex]{0pt}{0pt}\\
\cline{2-6}
% \rule{0pt}{2.5ex}& Gyawali \textit{et al.} \cite{gyawali2020semi} & Lung; Skin& Mixup on Input and Latent Space & Chexpert; ISIC 2018 &Chexpert: AUC: 0.6847; ISIC: AUC: 0.9073 \rule[-1.2ex]{0pt}{0pt}\\
% \cline{2-6}
% \rule{0pt}{2.5ex}& Kamran \textit{et al.} \cite{kamran2021vtgan}  &  Retina   &     Semi-supervised GAN + ViT &  FFA  &   FID (Fréchet inception distance): 17.3; KID (Kernel inception distance):0.00053 \rule[-1.2ex]{0pt}{0pt}\\
% \cline{2-6}
\rule{0pt}{2.5ex}& Yu \textit{et al.} \cite{yu2021accurate} & Colon & Mean Teacher  & Private Dataset: 13,111 Images &Nature Commun. 2021\rule[-1.2ex]{0pt}{0pt}\\
\bottomrule
\end{tabular}
\end{threeparttable}}
\label{tab:semi}
\end{table*}
\begin{table*}[ht]
\centering
\caption{Overview of Multi-instance Learning-based Studies in Medical Image Analysis.}

\resizebox{\textwidth}{.45\textwidth}{
\begin{threeparttable}
\begin{tabular}{@{}llllll@{}}
\toprule
 &Reference & Organ & MIL Algorithm Design & Dataset & Publication \\ \midrule
\rule{0pt}{2ex} \multirow{46}{*}{\rotatebox{90}{Classification}}&  Manivannan \textit{et al.} \cite{manivannan2017subcategory}$_{2017}$ &  Retina;   & Discriminative Subspace Transformation +  & Messidor; TMA-UCSB;  &TMI 2017\\
&&Breast&Margin-based Loss&DR Dataset; Private Dataset: 884 Images&\rule[-1.2ex]{0pt}{0pt}\\
\cline{2-6}
% \rule{0pt}{3ex}&Zhu \textit{et al.} \cite{zhu2017deep}$_{2017}$ & Breast & Sparse MIL & INBreast & AUC: 0.89\rule[-1.2ex]{0pt}{0pt}\\
% \cline{2-6}
% \rule{0pt}{3ex}& \rule{0pt}{2.5ex}Mercan \textit{et al.} \cite{mercan2017multi}$_{2017}$& Breast & Multi-Label MIL & BCSC & Average-P (Average-Precision): 0.8068\rule[-1.2ex]{0pt}{0pt}\\
% \cline{2-6}
\rule{0pt}{2.5ex}&Ilse \textit{et al.} \cite{ilse2018attention}$_{2017}$ &  Breast; Colon & Attention-based MIL   &  TMA-UCSB; CRCHistoPhenotypes &ICML 2018\rule[-1.2ex]{0pt}{0pt}\\
\cline{2-6}
\rule{0pt}{2.5ex}&Couture \textit{et al.} \cite{couture2018multiple}$_{2018}$ &  Breast  & Quantile Function-based MIL   &  CBCS3  &MICCAI 2018\rule[-1.2ex]{0pt}{0pt}\\
\cline{2-6}
% \rule{0pt}{2.5ex}&Das \textit{et al.} \cite{das2018multiple}$_{2018}$ &  Breast &  Multiple Instance Pooling   & BreakHis & Acc: 0.8906\rule[-1.2ex]{0pt}{0pt}\\
% \cline{2-6}
\rule{0pt}{2.5ex}&Liu \textit{et al.} \cite{liu2018landmark}$_{2018}$  & Brain & Landmark-based MIL    & ADNI; MIRIAD & MIA 2018\rule[-1.2ex]{0pt}{0pt}\\
\cline{2-6}
\rule{0pt}{2.5ex}&Campanella \textit{et al.} \cite{campanella2019clinical} &  Prostate; Skin; Lymph  & MIL + RNN   &  Private Dataset: 44,732 Images  & Nature Med. 2019\rule[-1.2ex]{0pt}{0pt}\\
\cline{2-6}
\rule{0pt}{2.5ex}&Wang \textit{et al.} \cite{wang2019rmdl} &  Breast &  Instance Features Recalibration   & Private Dataset: 608 Images & MIA 2019\rule[-1.2ex]{0pt}{0pt}\\
\bottomrule
\end{tabular}
\begin{tablenotes}    
        \footnotesize               
        \item[1] For the sake of brevity, we denote references that contain more than one task in the following abbreviations: \textbf{C}: Classification, \textbf{S}:Segmentation, \textbf{D}:Detection. 
      \end{tablenotes}
\end{threeparttable}
}
\label{tab:mil}
\end{table*}

\begin{table*}
\centering
\caption{Overview of Annotation-Efficient Learning Studies in Medical Image Analysis.}\label{tab:anno}
\resizebox{\textwidth}{.4\textwidth}{
\begin{threeparttable}
\begin{tabular}{@{}llllll@{}}
\toprule
& Reference & Organ & Method & Dataset & Publication\\
\midrule
\rule{0pt}{2.5ex} \multirow{8}{*}{\rotatebox{90}{Detection}} &\rule{0pt}{2.5ex}Hwang \textit{et al.}~\cite{hwang2016self}                                                     & Lung; Breast                                  & CAM + Self-Transfer Learning                                & Private Dataset: 11K X-rays; DDSM; MIAS 
 & MICCAI 2016\rule[-1ex]{0pt}{0pt}\\
\cline{2-6}

\rule{0pt}{2.5ex} &Gondal \textit{et al.}~\cite{gondal2017weakly}                                                 & Eye                                           & CAM                                                        & DRD; DiaretDB1  & ICIP 2017  \rule[-1ex]{0pt}{0pt} \\
\cline{2-6}

\rule{0pt}{2.5ex} &Wang \textit{et al.}~\cite{wang2018weakly}                                                    & Eye                                           & Expectation-Maximization  & DRD; Messidor & TMI 2018 \rule[-1ex]{0pt}{0pt}\\
\cline{2-6}

\rule{0pt}{2.5ex} &Wang \textit{et al.}~\cite{wang2020weakly} & Lung                                          & CAM + Unsupervised Segmentation                             & Private Dataset: 540 CT Images                                          & TMI 2020 \rule[-1ex]{0pt}{0pt}\\
\cline{2-6}

\rule{0pt}{2.5ex} &Shen \textit{et al.}~\cite{shen2021interpretable} & Breast                                        & Globally-aware Multiple Instance Classifier  & NYUBCS; CBIS-DDSM   & MIA 2021 \rule[-1ex]{0pt}{0pt}\\
\cline{2-6}

\rule{0pt}{2.5ex} & Liu~\textit{et al.}~\cite{liu_tssk-net_2023} & Eye & contrastive learning; knowledge distillation & Private: 7,000 OCT & CBM 2023 \\
\cline{2-6}

\rule{0pt}{2.5ex} & Wang~\textit{et al.}~\cite{wang2024weakly}  & Breast  & Two-Stage Detection and Diagnosis Network  & BUSI; BLUI  & ISBI 2024  \rule[-1ex]{0pt}{0pt} \\
\bottomrule
\end{tabular}
\end{threeparttable}
}
\end{table*}

\begin{table*}[ht]
\centering
\caption{Overview of Active Learning-based Studies in Medical Image Analysis.}

\resizebox{\textwidth}{.32\textwidth}{
\begin{threeparttable}
\begin{tabular}{@{}llllll@{}}
\toprule
 &Reference & Organ & Sampling Method & Dataset & Publication \\ \midrule
\rule{0pt}{2ex} \multirow{10}{*}{\rotatebox{90}{Classification}}&  Gal \textit{et al.} \cite{gal2017deep} &  Skin  & BALD + KL-divergence  &  ISIC 2016 & ICML 2017\rule[-1.5ex]{0pt}{0pt}\\ 
\cline{2-6}
\rule{0pt}{2.5ex}&  Wu \textit{et al.} \cite{wu2021covid} &  Lung  & Loss Prediction Network  & CC-CCII Dataset & MIA 2021\rule[-1.5ex]{0pt}{0pt}\\ 
\cline{2-6}
\rule{0pt}{3ex}& Li \textit{et al.} \cite{li2021pathal}  &  Prostate  & CurriculumNet + O2U-Net  &  ISIC 2017; PANDA Dataset & TMI 2021\rule[-1.5ex]{0pt}{0pt}\\
\cline{2-6}
\rule{0pt}{2.5ex}&  Qu \textit{et al.} \cite{qu2023openal} &  Colon  & Mahalanobis Distance  & NCT-CRC-HE-100K Dataset & MICCAI 2023\rule[-1.5ex]{0pt}{0pt}\\ 
\cline{2-6}
\rule{0pt}{2.5ex}&  Hu \textit{et al.} \cite{hu2023learning} & Colon; Lung & Gradient + Uncertainty & NCT-CRC-HE-100K Dataset; LC25000 & TMI 2023\rule[-1.5ex]{0pt}{0pt}\\ 
\cline{2-6}
\rule{0pt}{2.5ex}& \multirow{2}{*}{Mahapatra \textit{et al.} \cite{mahapatra2024gandalf}} &  \multirow{2}{*}{Chest}  & Graph Informativeness + & \multirow{2}{*}{CheXpert; ChestXray14; MedMNIST} & \multirow{2}{*}{MIA 2024}\rule[-1.5ex]{0pt}{0pt}\\
&&& Informative Augmentation && \\
\bottomrule
\end{tabular}
\begin{tablenotes}    
        \footnotesize               
        \item[1] For the sake of brevity, we denote references that contain more than one task in the following abbreviations: \textbf{C}: Classification, \textbf{S}: Segmentation, \textbf{D}: Detection. 
      \end{tablenotes}
\end{threeparttable}
}
\label{tab:al}
\end{table*}

\clearpage

\subsection{Datasets}
\begin{table*}[h]
\centering
\begin{center}
\small
\caption*{Appendix Table 6. Summary of publicly available databases for label-efficient learning in MIA.}
%\resizebox{\textwidth}{.64\textwidth}{
\begin{tabular}{@{}c|p{4cm}lp{0.39\textwidth}@{}}
% \begin{tabular}{@{}p{0.5cm}|p{3cm} p{2cm} p{9cm}@{}}
\toprule
\multicolumn{1}{c}{Organ}               & Dataset (Year)      & Task         & Link  \\ \hline
\multirow{32}{*}{Brain} & \rule{0pt}{2.5ex}BraTS (2012) & Segmentation & \url{http://www.imm.dtu.dk/projects/BRATS2012/data.html}   \\ 
\cline{2-4}
                       & \rule{0pt}{2.5ex}BraTS (2013)~\cite{menze2014multimodal}  & Segmentation & \url{https://www.smir.ch/BRATS/Start2013#!#download}   \rule[-1ex]{0pt}{0pt}\\
\cline{2-4}
                       & \rule{0pt}{2.5ex}BraTS (2015) & Segmentation & \url{https://www.smir.ch/BRATS/Start2015}   \rule[-1ex]{0pt}{0pt}\\
\cline{2-4}
                       & \rule{0pt}{2.5ex}BraTS (2017) & Segmentation & \url{https://sites.google.com/site/braintumorsegmentation/}   \rule[-1ex]{0pt}{0pt}\\
\cline{2-4}
                       & \rule{0pt}{2.5ex}BraTS (2018) & Segmentation & \url{https://wiki.cancerimagingarchive.net/pages/viewpage.action?pageId=37224922}   \rule[-1ex]{0pt}{0pt}\\
\cline{2-4}
                       & \rule{0pt}{2.5ex}MSD (2018)\cite{simpson2019large} &Segmentation &\url{https://drive.google.com/drive/folders/1HqEgzS8BV2c7xYNrZdEAnrHk7osJJ--2}   \rule[-1ex]{0pt}{0pt}\\
\cline{2-4}
                       & \rule{0pt}{2.5ex}dHCP (2018) \cite{makropoulos2018developing}            &Segmentation     &\url{http://www.developingconnectome.org/data-release/}       \rule[-1ex]{0pt}{0pt}\\
\hline
\end{tabular}%}
\label{tab:Dataset1}
% \end{tabular}
\end{center}
\end{table*}

\begin{table*}[h]
\small
\centering
\begin{center}
\caption*{Appendix Table 6. Summary of publicly available databases for label-efficient learning in MIA. (continued)}
\begin{tabular}{@{}c|p{4cm}lp{0.4\textwidth}@{}}
\toprule
\multicolumn{1}{c}{Organ}               & Dataset (Year)      & Task         & Link  \\ \hline
\multirow{23}{*}{Chest (Continued)} 
                &\rule{0pt}{2.5ex}CAD-PE (2019)              &Segmentation              &\url{https://ieee-dataport.org/open-access/cad-pe}       \rule[-1ex]{0pt}{0pt}\\
\cline{2-4}
                       &\rule{0pt}{2.5ex}SIIM-ACR (2019)              &Segmentation              &\url{https://www.kaggle.com/c/siim-acr-pneumothorax-segmentation}       \rule[-1ex]{0pt}{0pt}\\
\cline{2-4}
                       &\rule{0pt}{2.5ex}RSNA-Lung (2018)              &Detection              &\url{https://www.kaggle.com/c/rsna-pneumonia-detection-challenge}       \rule[-1ex]{0pt}{0pt}\\
\cline{2-4}
                       &\rule{0pt}{2.5ex}VinDr-CXR (2021)              &Detection              &\url{https://vindr.ai/datasets/cxr}       \rule[-1ex]{0pt}{0pt}\\
\cline{2-4}
                       &\rule{0pt}{2.5ex}Montgomery (2022)              &Segmentation              &\url{https://www.kaggle.com/datasets/raddar/tuberculosis-chest-xrays-montgomery}       \rule[-1ex]{0pt}{0pt}\\
% \cline{2-4}
%                        &\rule{0pt}{2.5ex}RICORD (2021)~\cite{tsai2021rsna}              &Segmentation              &\url{https://wiki.cancerimagingarchive.net/pages/viewpage.action?pageId=80969742}       \rule[-1ex]{0pt}{0pt}\\
\cline{2-4}
                       &\rule{0pt}{2.5ex}ChestXR (2021)              &Classification              &\url{https://cxr-covid19.grand-challenge.org/Dataset/}       \rule[-1ex]{0pt}{0pt}\\
\cline{2-4}
                       &\rule{0pt}{2.5ex}MIMIC-CXR (2019)~\cite{johnson2019mimic}             &Detection              &\url{https://physionet.org/content/mimic-cxr/2.0.0/}       \rule[-1ex]{0pt}{0pt}\\
\hline
\end{tabular}\label{tab:Dataset2}
\end{center}
\end{table*}

\begin{table*}[h]
\small
\centering
\begin{center}
\caption*{Appendix Table 6. Summary of publicly available databases for label-efficient learning in MIA. (continued)}

\begin{tabular}{@{}c|p{4cm}lp{0.4\textwidth}@{}}
% \begin{tabular}{@{}p{0.5cm}|p{3cm} p{2cm} p{9cm}@{}}
\toprule
\multicolumn{1}{c}{Organ}               & Dataset (Year)      & Task         & Link  \\ \hline
\multirow{28}{*}{Eye (Continued)}
                       & \rule{0pt}{2.5ex}REFUGE~\cite{orlando2020refuge}& Segmentation& \url{https://ieee-dataport.org/documents/refuge-retinal-fundus-glaucoma-challenge}\rule[-1ex]{0pt}{0pt}\\
\cline{2-4}
                       & \rule{0pt}{2.5ex}DRD (2015)& Detection  & \url{https://www.kaggle.com/competitions/diabetic-retinopathy-detection}\rule[-1ex]{0pt}{0pt}\\
\cline{2-4}
                       & \rule{0pt}{2.5ex}RetinalOCT (2018)~\cite{kermany2018identifying}& Classification & \url{https://www.kaggle.com/datasets/paultimothymooney/kermany2018}\rule[-1ex]{0pt}{0pt}\\
% \cline{2-4}
%                        & \rule{0pt}{2.5ex}DiaretDB1 (2007)~\cite{kauppi2007diaretdb1}& Detection & \url{https://www.it.lut.fi/project/imageret/diaretdb1/}\rule[-1ex]{0pt}{0pt}\\
\cline{2-4}
                       & \rule{0pt}{2.5ex}ReTOUCH (2017)~\cite{bogunovic2019retouch} & Classification & \url{https://retouch.grand-challenge.org/}\rule[-1ex]{0pt}{0pt}\\
\cline{2-4}
                       & \rule{0pt}{2.5ex}ORIGA-light (2010)~\cite{zhang2010origa} & Classification & \url{https://www.medicmind.tech/retinal-image-databases}  \rule[-1ex]{0pt}{0pt}\\
\cline{2-4}
                       & \rule{0pt}{2.5ex}sjchoi86-HRF (2017) & Classification & \url{https://github.com/cvblab/retina_dataset} \rule[-1ex]{0pt}{0pt}\\
\hline
\end{tabular}\label{tab:Dataset3}
\end{center}
\end{table*}

\begin{table*}[h]
\small
\centering
\begin{center}
\caption*{Appendix Table 6. Summary of publicly available databases for label-efficient learning in MIA. (continued)}

\begin{tabular}{@{}c|p{4cm}lp{0.4\textwidth}@{}}
% \begin{tabular}{@{}p{0.5cm}|p{3cm} p{2cm} p{9cm}@{}}
\toprule
\multicolumn{1}{c}{Organ}               & Dataset (Year)      & Task         & Link  \\ \hline
\multirow{3}{*}{Abdomen} & \rule{0pt}{2.5ex}ACDC (2018)~\cite{bernard2018deep}& Segmentation  & \url{https://www.creatis.insa-lyon.fr/Challenge/acdc/databases.html}   \rule[-1ex]{0pt}{0pt}\\
\cline{2-4}
                       & \rule{0pt}{2.5ex}CHAOS (2021)~\cite{kavur2021chaos}& Segmentation & \url{https://chaos.grand-challenge.org/} \rule[-1ex]{0pt}{0pt}\\
\hline
\multirow{28}{*}{Breast} & \rule{0pt}{2.5ex}BACH (2018) \cite{aresta2019bach} &    Classification & \url{https://iciar2018-challenge.grand-challenge.org/Dataset/}   \\ 
\cline{2-4}
                       &\rule{0pt}{2.5ex}NYUBCS (2019)~\cite{wu2019nyu}& Segmentation & \url{https://datacatalog.med.nyu.edu/dataset/10518}  \rule[-1ex]{0pt}{0pt}\\
\cline{2-4}
                       &\rule{0pt}{2.5ex}CBIS-DDSM (2017)~\cite{lee2017curated}& Segmentation & \url{https://www.kaggle.com/datasets/awsaf49/cbis-ddsm-breast-cancer-image-dataset}  \rule[-1ex]{0pt}{0pt}\\
% \cline{2-4}
%                        &\rule{0pt}{2.5ex}DDSM (1998)~\cite{heath1998current}& Detection & \url{https://www.kaggle.com/datasets/skooch/ddsm-mammography} \rule[-1ex]{0pt}{0pt}\\
\cline{2-4}
                       &\rule{0pt}{2.5ex}MIAS (2015)~\cite{suckling2015mammographic}& Detection & \url{https://www.kaggle.com/datasets/kmader/mias-mammography} \rule[-1ex]{0pt}{0pt}\\
\cline{2-4}
                       &\rule{0pt}{2.5ex}TCGA-Breast & Classification & \url{https://portal.gdc.cancer.gov/repository} \rule[-1ex]{0pt}{0pt}\\
\hline
\end{tabular}\label{tab:Dataset3-2}
\end{center}
\end{table*}

\begin{table*}[h]
\small
\centering
\begin{center}
\caption*{Appendix Table 6. Summary of publicly available databases for label-efficient learning in MIA. (continued)}

\begin{tabular}{@{}c|p{4cm}lp{0.4\textwidth}@{}}
% \begin{tabular}{@{}p{0.5cm}|p{3cm} p{2cm} p{9cm}@{}}
\toprule
\multicolumn{1}{c}{Organ}               & Dataset (Year)      & Task         & Link  \\ \hline
\multirow{18}{*}{Multi-organ} & \rule{0pt}{2.5ex}DSB (2018)& Segmentation & \url{https://www.kaggle.com/competitions/data-science-bowl-2018/data}  \\
\cline{2-4}
                       &\rule{0pt}{2.5ex}DeepLesion (2018)~\cite{yan2018deeplesion}& Detection & \url{https://nihcc.app.box.com/v/DeepLesion} \rule[-1ex]{0pt}{0pt}\\
\cline{2-4}
                       &\rule{0pt}{2.5ex}WTS (2020)~\cite{keikhosravi2020non}& Super-resolution & \url{https://www.nature.com/articles/s42003-020-01151-5#data-availability} \rule[-1ex]{0pt}{0pt}\\
                       
\cline{2-4}
                       &\rule{0pt}{2.5ex}DECATHLON (2019)~\cite{simpson2019large}& Segmentation & \url{http://medicaldecathlon.com/} \rule[-1ex]{0pt}{0pt}\\
\cline{2-4}
                       &\rule{0pt}{2.5ex}MoNuSeg (2017)~\cite{kumar2017dataset}& Segmentation & \url{https://monuseg.grand-challenge.org/} \rule[-1ex]{0pt}{0pt}\\
\cline{2-4}
                       &\rule{0pt}{2.5ex}MoCTSeg (2018)~\cite{gibson2018automatic}& Segmentation & \url{https://www.synapse.org/#!Synapse:syn3376386} \rule[-1ex]{0pt}{0pt}\\
\cline{2-4}
                       &\rule{0pt}{2.5ex}BTCV (2017)~\cite{gibson2018multi}& Segmentation & \url{https://zenodo.org/record/1169361#.Y8Ud-OxBwUE} \rule[-1ex]{0pt}{0pt}\\
\hline
\end{tabular}\label{tab:Dataset3-3}
\end{center}
\end{table*}
\end{document}